\documentclass{article}

\usepackage[final]{neurips_2022}

\usepackage{titletoc}
\usepackage[utf8]{inputenc} % allow utf-8 input
\usepackage[T1]{fontenc}    % use 8-bit T1 fonts
\usepackage{hyperref}       % hyperlinks
\usepackage{url}            % simple URL typesetting
\usepackage{booktabs}       % professional-quality tables
\usepackage{amsfonts}       % blackboard math symbols
\usepackage{amsthm}
\usepackage{nicefrac}       % compact symbols for 1/2, etc.
\usepackage{microtype}      % microtypography
\usepackage{xcolor}         % colors
\usepackage{subcaption}
\usepackage{graphicx}
\usepackage{amsmath}
\usepackage{multirow}
\usepackage{wrapfig}
\usepackage{tikz}
\usepackage{amssymb}
\usepackage{pifont}% http://ctan.org/pkg/pifont
\usepackage[colorinlistoftodos]{todonotes}
\usepackage[normalem]{ulem} 
\usepackage{caption}
\usepackage{listings}
\usepackage[page,header]{appendix}

\usepackage{arydshln}

\theoremstyle{definition}
\newtheorem{definition}{Definition}[section]
\newcommand{\cmark}{\ding{51}}%
\newcommand{\xmark}{\ding{55}}%

\definecolor{our_purple}{HTML}{8C5DEC}
\definecolor{our_green}{HTML}{0FA3B1}
\definecolor{our_maroon}{HTML}{F17C81}

\title{Learning to Follow Instructions in Text-Based Games}
\author{
Mathieu Tuli, Andrew C. Li, Pashootan Vaezipoor, Toryn Q. Klassen$^\dagger$,\\ {\bf Scott Sanner, Sheila A. McIlraith$^\dagger$}\\
 University of Toronto, Toronto, Canada \\ Vector Institute for Artificial Intelligence, Toronto, Canada\\ $^\dagger$ Schwartz Reisman Institute for Technology and Society, Toronto, Canada
 \\
 \texttt{\{mathieutuli,andrewli,pashootan,toryn,sheila\}@cs.toronto.edu}\\ \texttt{ssanner@mie.utoronto.ca}
 }

\definecolor{light-grey}{RGB}{120, 120, 120}
\definecolor{BrickRed}{RGB}{140, 0, 0}
\definecolor{LighterBlue}{RGB}{47, 140, 255}
\definecolor{lightblue}{RGB}{200, 200, 255}
\newif\ifcomments
\commentstrue
\ifcomments
\newcommand{\cut}[1]{}
\newcommand{\maybecut}[1]{\textcolor{orange}{#1}}
\newcommand{\debating}[1]{\textcolor{yellow}{#1}}

\newcommand{\donepromised}[1]{{}}
\newcommand{\toresolve}[1]{\textcolor{red}{#1}}
\newcommand{\remove}[1]{\textcolor{green}{\sout{#1}}}

\newcommand{\removehide}[1]{}
\newcommand{\alt}[1]{\textcolor{brown}{#1}}

\newcommand{\commentsm}[1]{\textcolor{cyan}{({\bf SM:} #1)}}

\newcommand{\commentpv}[1]{\textcolor{LighterBlue}{({\bf PV:} #1)}}
\newcommand{\commental}[1]{\textcolor{magenta}{({\bf AL:} #1)}}

\newcommand{\old}[1]{\textcolor{red}{\sout{#1}}}
\newcommand{\new}[1]{\textcolor{BrickRed}{\uline{#1}}}
\else
    \newcommand{\commentsm}[1]{}
    \newcommand{\commentrti}[1]{}
    \newcommand{\commentpv}[1]{}
    \newcommand{\commental}[1]{}
    \newcommand{\maybecut}[1]{}
    \newcommand{\debating}[1]{}
    
    \newcommand{\donepromised}[1]{}
    \newcommand{\toresolve}[1]{}
    \newcommand{\remove}[1]{}
    \newcommand{\removehide}[1]{}
    \newcommand{\alt}[1]{}
    \newcommand{\old}[1]{}
    \newcommand{\new}[1]{}

\fi

\definecolor{light-grey}{RGB}{120, 120, 120}

\usepackage{mathtools}
\usepackage{amssymb}
\usepackage{amsmath}
\usepackage{xspace}

\newcommand{\RLTL}{R_\text{LTL}}
\newcommand{\codeurl}{\url{https://github.com/MathieuTuli/LTL-GATA}}

\newcommand{\expect}{\mathbb{E}}

\newcommand{\tuple}[1]{\langle{#1}\rangle}

\DeclareMathOperator*{\argmax}{arg\,max}

\newcommand{\ltlalways}{\ensuremath{\Box}\xspace} 
\newcommand{\ltleventually}{\ensuremath{\Diamond}\xspace}
\newcommand{\ltlnext}{\ensuremath{\bigcirc}\xspace}

\newcommand{\true}{\ensuremath{\mathsf{true}}\xspace{}}
\newcommand{\false}{\ensuremath{\mathsf{false}}\xspace{}}
\newcommand{\ltluntil}{\ensuremath{\operatorname{\mathsf{U}}}\xspace{}}

\newcommand{\textltluntil}{\ensuremath{\operatorname{\mathsf{U}}}\xspace}

\newcommand{\mprog}{\operatorname{prog}}

\newcommand{\ltlalwaysTXT}{\ensuremath{\operatorname{\texttt{ALWAYS}}}}
\newcommand{\ltleventuallyTXT}{\ensuremath{\operatorname{\texttt{EVENTUALLY}}}}
\newcommand{\ltlnextTXT}{\ensuremath{\operatorname{\texttt{NEXT}}}}

\newcommand{\ltluntilTXT}{\ensuremath{\operatorname{\texttt{UNTIL}}}}
\newcommand{\ltlpredicate}[1]{\ensuremath{\texttt{#1}}}

\newcommand{\agentname}{LTL-GATA}

\begin{document}
\newpage
% we perhaps just want to uncomment this skeleton, the cover page doesn't hold much atm
% \input{skeleton}
\newpage

\maketitle

\begin{abstract}
     Text-based games present a unique class of sequential decision making problem in which agents interact with a partially observable, simulated environment via actions and observations conveyed through
     %that are stated in 
     natural language.
     %
     %Alternative using observations and actions that are communicated via natural language. 
     % Alternative:  ... simulated environment via natural language observations and actions.
     %
     Such observations typically include instructions that, in a reinforcement learning (RL) setting, can directly or indirectly guide a player towards completing reward-worthy tasks. 
     In this work, we study the ability of RL agents to follow such instructions. We conduct experiments that show that the performance of state-of-the-art text-based game agents is largely unaffected by the presence or absence of such instructions, and that these agents are typically unable to execute tasks to completion.
     %, and that learned behaviors sometimes include \revisit{extraneous or detrimental actions}. 
     %
     %suggesting that these RL agents are not benefiting from these instructions.
     %
     To further study and address the task of instruction following, we equip RL agents with an internal structured representation of  natural language instructions in the form of Linear Temporal Logic (LTL), a formal language that is increasingly used for temporally extended reward specification in RL. Our framework both supports and highlights the benefit of understanding the temporal semantics of instructions and in measuring progress towards achievement of such a temporally extended behaviour. %, such as a set of instructions.
     Experiments with 500+ games in TextWorld demonstrate the superior performance of our approach.

\end{abstract}
\section{Introduction}
\label{sec:introduction}

Building AI agents that can understand natural language is an important and longstanding problem in AI. In recent years, instrumented text-based game (TBG) engines have served as compelling environments for studying a variety of tasks related to language understanding, affordance extraction,
memory, and sequential decision making \citep[e.g.,][]{cote2018textworld, adhikari2020learning, liu2022learning}.  They provide a simulated, partially observable environment where an agent can navigate and interact with environment objects, receiving observations and administering commands via natural language.  
\begin{figure}[t!]
    \centering
    \includegraphics[width=0.9\textwidth]{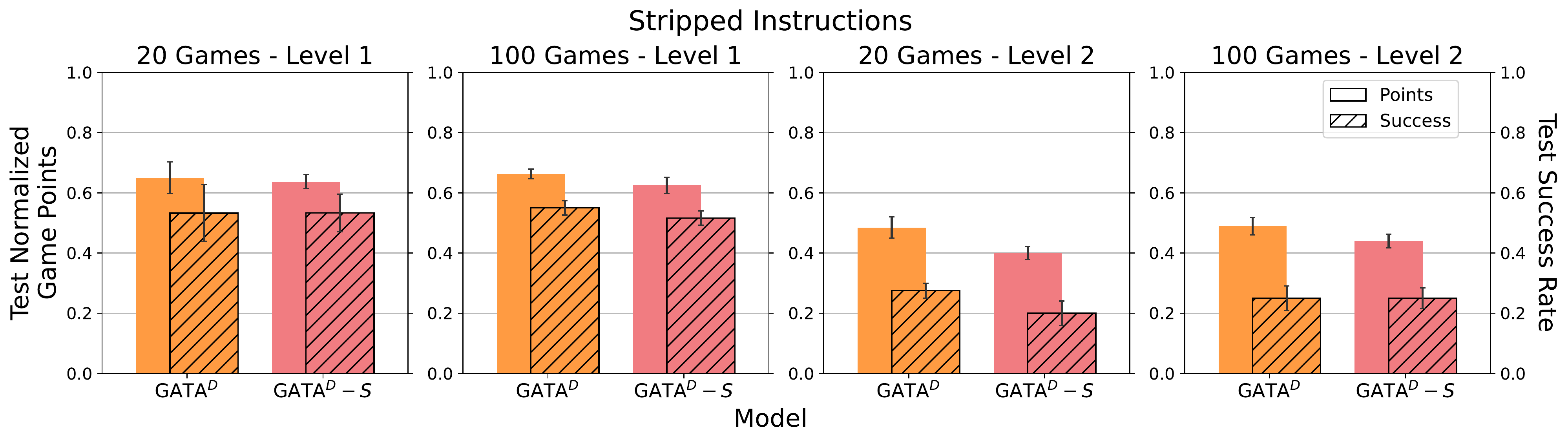}
    \caption{Comparison of GATA performance when trained with instructions (GATA\textsuperscript{D}) versus when instructions are stripped from environment observations (GATA\textsuperscript{D}-S). Agents were trained with 20 or 100 games, at increasing levels of task difficulty (level 1 vs level 2). Note that normalized game point performance (solid blocks) and rate of success (hashed blocks) are largely unchanged whether instructions are present or absent. Low success rate (i.e., task completion) rate is also seen in level 2.}
    %GATA\textsuperscript{D} represents the model trained using normal observations, and GATA\textsuperscript{D}-S is the model trained on the \emph{stripped} observations. 
    %\remove{GATA\textsuperscript{D}-S therefore does not capture goal relations like `needs', 
    %but despite not having instructions is equally performant to GATA\textsuperscript{D}. 
    %The solid bars show normalized game points for each approach, while the hatched bars show success rate -- task completion, both averaged across the same 20 test games. The first and third graphs show results when the agents are trained on 20 games unique, and the second and fourth when the agent are trained on 100 games unique.}
%    }
    \label{fig:stripped}
\end{figure}
TextWorld \citep{cote2018textworld} is a TBG learning environment for training reinforcement learning (RL) agents.  Successful play requires language understanding, effective navigation, memory, and an ability to follow instructions embedded within the text. 
Instructions may or may not be directly bound to reward but can guide an RL agent towards completing tasks and collecting reward.

% i don't think the claim is too strong nor incorrect, we do do this, and it's even true for ICLR even if we don't beat them on all levels or discuss it, so I think it's okay.
In this paper we study instruction following in text-based games and propose an approach that advances the previous state of the art. To this end, we employ the state-of-the-art model-free TBG RL agent called GATA (Graph Aided Transformer Agent) \citep{adhikari2020learning} that operates in the TextWorld environment. GATA has made significant advances in performance by augmenting TBG agents with long-term memory – a critical component of effective game play. Despite GATA's improvement over previous baselines, our experiments (see \autoref{fig:stripped}) show that GATA performance is largely unaffected by the presence or absence of instructions, leading us to conclude that GATA is not effectively following instructions. We also find that while GATA agents are able to garner reward, they are not typically successful in \emph{completing} tasks – an important vulnerability to the deployment of such techniques in environments where partial completion of tasks can be unsafe.
%Intuitively, without knowing the goal or following instructions to achieve the goal, agents couldn't reasonably succeed in solving these games. 
%\removemaybe{This is particularly apparent in the on Level 2 experiements in \autoref{fig:stripped}, where the success rate of games is roughly half that of the achieved normalized game points.}

To further study and address the task of instruction following, we equip GATA with an internal structured representation of natural language instructions specified in Linear Temporal Logic (LTL) \citep{DBLP:conf/focs/Pnueli77}, a formal language that is increasingly used for temporally extended goals in planning 
and reward specification and other purposes 
% \commenttk{I can't say ``advice'', since it's not really ``increasingly used'' for advice.}
in RL \citep[e.g.,][]{
% Planning:
DBLP:journals/ai/BacchusK00,
bai-mci-icaps06,
DBLP:conf/ijcai/PatriziLGG11,
cam-mci-ijcai19,
% RL:
DBLP:journals/corr/LittmanTFIWM17, 
icarte2018advice, % not really reward specification
% hasanbeig2018logically,
toro2018teaching, camacho-etal-ijcai19, 
% DBLP:conf/kr/GiacomoFIPR20, 
leon2020systematic, kuo2020encoding, vaezipoor2021ltl2action}. 
%\commentsm{Toryn could you please add a bunch of citations here with an "e.g.,".  We should cite the AAMAS paper, LTL2Action, and any other published works of ours and others, including the advice paper from Cdn AI} 
% \commenttk{The advice paper isn't really an example of LTL reward specification.} \commentsm{I'd rephrase so it's included. It's an important example of the use of LTL in RL and I think it could potentially be used for the navigation work. -- simple rephrase of the sentence that prompts the citations.  Thx.}
% We likewise enhance the GATA architecture with 
%a \comment{graph-based encoding}{TK: Didn't GATA already have a graph encoder? SM: yes. Not sure whehter our LTL2Action one for the LTL is different. Mat can you clarify? Thx MT: it's not different} and 
LTL also provides a mechanism to monitor progress towards completion of instructions. Our framework both supports and highlights the benefit of understanding the temporal semantics of instructions and in measuring progress towards achievement of a temporally extended behaviour. We perform experiments that illustrate the superior performance of our TBG agent and its ability to follow instructions. Contributions of this work include: 
\begin{itemize}
    \item Experiments that expose the lack of instruction following and low task completion rate in a state-of-the-art TBG agent.
    %, despite relatively good point-measured performance. %
    % I would avoid the above, commented out part.
    \item An approach to the study and deployment of instruction following in TBG environments via exploitation of a formal language: LTL. LTL provides well-defined semantics and supports a measure of progress towards satisfaction of instructions.
    \item An augmentation to an existing state-of-the-art architecture for TBGs to equip a TBG agent with instruction-following capabilities.
    \item Comprehensive experiments and insights that study our and others' approaches to instruction following, and that highlight the superior performance of our proposed approach.
\end{itemize}

%\commentmt{please comment/amend}
% \commenttk{I assume these bullet points aren't finished.}\commentsm{Correct. I wanted some input from others, if possible.  Any thoughts?}
% \commenttk{I wonder if we need an explicit list of contributions. They seem to be somewhat repeating what the introduction has already said. But perhaps we need them to avoid reviewers underweighting the analysis of GATA's limitations?}
% \commentmt{I like having them for reviwers that like to skim.}
% \input{sections/introduction}
\section{Background}
\label{sec:background}

In this section we introduce TextWorld, the TBG engine that we use, together with the Cooking domain that we employ in our experiments. We also overview Linear Temporal Logic, which (as described in \autoref{sec:introduction}) we use in our approach as an internal representation for instructions.

\subsection{Text-Based Games: TextWorld}
\label{sec:text_based_games}
Text-based games are partially observable multi-turn games where the environment and the player's action choices are represented textually. In this work, we use TextWorld \citep{cote2018textworld} as our text-based game engine. A text-based game can be viewed as a (discrete-time) partially observable Markov decision process (POMDP) $\tuple{S, T, A, O, \Omega, R, \gamma}$ \citep{cote2018textworld} where $S$ is the environment's state space, $A$ is the action space, $T(s_{t+1} | s_t, a_t)$ where $s_{t+1}, s_t \in S$ and $a_t \in A$ is the conditional transition probability between states $s_{t+1}$ and $s_t$ given action $a_t$, $O$ is the set of (partial) observations that the agent receives, $\Omega(o_t | s_t, a_{t-1})$ is the set of conditional observation probabilities, $R : S \times A \rightarrow \mathbb{R}$ is the reward function, and $\gamma \in [0, 1]$ is the discount factor. An agent's goal is to learn some optimal policy $\pi^*(a | o)$ (or a policy that conditions on historical observations or on some internal memory) that maximizes the expected discounted return.
% Text-based games are played one of three ways: (1) choice-based games where an agent selects from a choice of actions at each step in an episode: (2) hypertext-based games where an agents clicks on one of several links from the current text observations: (3) parser-based games where agents must type text commands chosen from a predefined vocabulary to compose commands \citep{cote2018textworld}. 
In this work, we focus on the choice-based variant of games, similar to previous works \citep{adhikari2020learning, narasimhan2015language}. The action space $A$ is a list of possible commands and at each time-step $t$ in the game, the agent must select action $a_t \in C_t$ from the current subset of permissible actions $C_t \subset A$. 
%\commentmt{should we ref. the data-flow figure (fig 2) here?}. 
%\commenttk{I think that shows too many other things as well.}

\subsubsection{Environment Setting}
% \commentmt{@Toryn, the ICLR work cited the original textworld paper, GATA cited no one and said "similar" like me, the problem is the original TW paper only talks about coin collection game, and the cooking game is part of their "benchmarks" but it's not mentioend in the original paper. It can be found in their python docs though}
% \commentmt{the cooking domain was first used in the "first textworld problems", GATA just generated and compiled a serperate instance of games using the updated textworld engine at the time.}
% We focus on the popular TextWorld \textit{Cooking domain} \comment{from \citet{adhikari2020learning}}{I think the reason GATA cited no-one for the cooking domain is that they themselves introduced it.} (which is similar to
% %TK: Is there somewhere we can cite for the game itself, instead of just something it's similar to?
% the games used in Microsoft's \textit{First TextWorld Problems: A Language and Reinforcement Learning Challenge} (FTWP) \citep{trischlerBlog2019first}), 
We focus on the TextWorld \textit{Cooking domain}, popularized by \cite{adhikari2020learning} and Microsoft's First TextWorld Problems: A Language and Reinforcement Learning Challenge (FTWP) \citep{trischlerBlog2019first}. The game tasks agents with gathering and preparing various cooking ingredients described by an in-game recipe that is to be found. 
% Game points (rewards) are earned by \comment{\remove{some combination}}{TK: This is a bit vague. Points can be earned for each of them, right? (Provided it's in the recipe.) It's not that, e.g., the agent has to do both (1) and (2) to get any reward.} \added{each} of (1) collecting the proper ingredients, (2) preparing them\added{, if required by the recipe} (preparation can involve some cutting action and/or some cooking action), (3) preparing the meal once all of the ingredients have been prepared, and (4) eating the meal. 
Game points (rewards) are earned for each of (1)~collecting a required ingredient, 
%(2) preparing an ingredient in a way required by the recipe (preparation can involve some cutting action \comment{and/or}{Is there a separate reward for each step of preparation? I'm assuming so. (mat: yes)} some cooking action: reward is assigned for each action), 
(2) performing a preparatory step (some cutting or cooking action) on an ingredient as required by the recipe,
(3) preparing the meal once all of the ingredients have been prepared, and (4) eating the meal.
The game's partial observations can contain instructions that guide the agent towards completion of tasks, but not all instructions correspond directly to rewards. The game first instructs the agent to examine a cookbook, which elicits a recipe to be followed. The act of examining the cookbook returns no reward, but following its recipe will return reward. See \autoref{sec:app_tw} for more details. Success is determined by whether the recipe is fully completed and eaten. Preparing ingredients can also involve collecting certain tools (e.g., a knife). The game may also involve navigation -- the agent may need to navigate to the kitchen or to find certain ingredients.

\subsection{Linear Temporal Logic (LTL)}
\label{sec:ltl}
% \citep{DBLP:conf/focs/Pnueli77} - to be used when first mentioning LTL
Linear Temporal Logic (LTL) \citep{DBLP:conf/focs/Pnueli77} is a formal language -- a propositional logical language with temporal modalities -- that can be used to describe properties of trajectories. We will use LTL to specify instructions. LTL formulas are constructed from propositional variables (e.g., \ltlpredicate{player-has-carrot}), connectives from propositional logic (e.g. $\neg$), and two temporal operators: $\ltlnext$ ($\ltlnextTXT$) and $\textltluntil$ ($\ltluntilTXT$). Formally, we define the syntax of LTL per \cite{DBLP:books/daglib/0020348}
%\cite{DBLP:books/daglib/0020348} <- should we still include ?
as
\begin{equation*}
    \varphi \Coloneqq p \;|\; \neg \varphi \;|\; \varphi \wedge \psi \;|\; \ltlnext \varphi \;|\; \varphi \ltluntil \psi
\end{equation*}
where $p \in \mathcal{P}$ for some finite set of propositional symbols $\mathcal{P}$.
Satisfaction of an LTL formula is determined by a sequence of truth assignments $\sigma=\langle \sigma_0, \sigma_1, \sigma_2, \ldots \rangle$ for $\mathcal{P}$, where $p \in \sigma_i$ iff proposition $p \in \mathcal{P}$ holds at time step $i$. Formally, $\sigma$ \emph{satisfies} $\varphi$ at time $i \geq 0$, denoted as $\tuple{\sigma,i}\models\varphi$, under the following conditions:

\begin{tabular}{ll}
    $\bullet\ \tuple{\sigma,i}\models p$ iff $p\in \sigma_i$, where $p \in \mathcal{P}$
    &
    $\bullet\ \tuple{\sigma,i}\models (\varphi\wedge\psi)$ iff $\tuple{\sigma,i}\models\varphi$ and $\tuple{\sigma,i}\models\psi$
    \\
    $\bullet\ \tuple{\sigma,i}\models \neg \varphi$ iff $\tuple{\sigma,i}\not\models\varphi$ 
    &
    $\bullet\ \tuple{\sigma,i}\models\varphi\ltluntil\psi$ iff there exists $j$ such that $i\le j$ %\le n$ % removed per Sheila's comment
        and
    \\
    $\bullet\ \tuple{\sigma,i}\models\ltlnext\varphi$ iff $\tuple{\sigma,i+1}\models\varphi$
    & 
        ${}\qquad\qquad\tuple{\sigma,j}\models\psi$, and $\tuple{\sigma,k}\models\varphi$ for all $k \in [i,j)$
\end{tabular}

A sequence $\sigma$ is then said to \emph{satisfy} $\varphi$ iff $\tuple{\sigma,0}\models\varphi$.

Any LTL formula can be defined in terms of $p \in \mathcal{P}$, $\neg$ (\emph{negation}), $\wedge$ (\emph{and}), $\ltlnext$ ($\ltlnextTXT$), and $\textltluntil$ ($\ltluntilTXT$). From these operators, we can also define the Boolean operators $\vee$ (\emph{or}) and $\rightarrow$ (\emph{implication}), and the temporal operators $\ltlalways$ ($\ltlalwaysTXT$) and $\ltleventually$ ($\ltleventuallyTXT$), where $\tuple{\sigma,0}\models\ltlalways \varphi$ if $\varphi$ always holds in $\sigma$, and $\tuple{\sigma,0}\models\ltleventually \varphi$ if $\varphi$ holds at some point in $\sigma$.

\subsubsection{LTL Progression}
\label{sec:prog}
LTL formulas can also be \textit{progressed} along a sequence of truth assignments \citep{DBLP:journals/ai/BacchusK00,toro2018teaching}.
% \comment{\remove{\citep{bacchus1996rewarding, toro2018teaching}}}{TK: \cite{bacchus1996rewarding} used past LTL, so used regression, not progression.} \added{\citep{DBLP:journals/ai/BacchusK00,toro2018teaching}}. 
% you're right, I had the wrong bacchus citation, thanks
In other words, as an agent acts in the environment, resulting truth assignments can be used to update the formula to reflect what has been satisfied. The updated formula would now reflect the parts of the original formula that are remaining to be satisfied or whether the formula has been violated/satisfied. The progression operator $\mprog(\sigma_i, \varphi)$ is defined as follows.
\begin{definition}
For LTL formula $\varphi$, truth assignment $\sigma_i$ over $\mathcal{P}$, and $p \in \mathcal{P}$, $\mprog(\sigma_i, \varphi)$ is defined as\\
\begin{tabular}{ll}
    % &\mprog(\sigma_i, p) = \true~\text{if}~p \in \sigma_i\\
    % &\mprog(\sigma_i, p) = \false~\text{if}~p \notin \sigma_i\\
    $\bullet\ \mprog(\sigma_i, p) = \begin{cases}\true&\text{if}~p \in \sigma_i\\\false&\text{otherwise}\end{cases}$&$\bullet\ \mprog(\sigma_i, \varphi_1 \wedge \varphi_2) = \mprog(\sigma_1, \varphi_1) \wedge \mprog(\sigma_1, \varphi_2)$\\
    
    $\bullet\ \mprog(\sigma_i, \neg\varphi) = \neg\mprog(\sigma_i, \varphi)$&$\bullet\ \mprog(\sigma_i, \varphi_1\ltluntilTXT\varphi_2) = $\\
    
    $\bullet\ \mprog(\sigma_i, \ltlnextTXT\varphi) = \varphi$& $\quad\quad\quad \mprog(\sigma_1, \varphi_2) \vee  (\mprog(\sigma_1, \varphi_1) \wedge \varphi_1 \ltluntilTXT \varphi_2)$\\
\end{tabular}

% \begin{tabular}{ll}
%     $\bullet\ \tuple{\sigma,i}\models p$ iff $p\in \sigma_i$, where $p \in \mathcal{P}$
%     &
%     $\bullet\ \tuple{\sigma,i}\models\varphi\ltluntil\psi$ iff there exists $j$ such that $i\le j$ %\le n$ % removed per Sheila's comment
%         and
%     \\
%     $\bullet\ \tuple{\sigma,i}\models \neg \varphi$ iff $\tuple{\sigma,i}\not\models\varphi$ 
%     &
%         ${}\qquad\qquad\tuple{\sigma,j}\models\psi$, and $\tuple{\sigma,k}\models\varphi$ for all $k \in [i,j)$
%     \\
%     $\bullet\ \tuple{\sigma,i}\models\ltlnext\varphi$ iff $\tuple{\sigma,i+1}\models\varphi$
%     & 
%     $\bullet\ \tuple{\sigma,i}\models (\varphi\wedge\psi)$ iff $\tuple{\sigma,i}\models\varphi$ and $\tuple{\sigma,i}\models\psi$
% \end{tabular}
\end{definition}
In the context of TextWorld, the progression operator can be applied at every step in the episode to update the LTL instruction fed to the agent. 
To do so, it's necessary to have \textit{event detectors} that can detect when propositions are true as the agent acts during an episode (e.g., to detect that \ltlpredicate{player-has-carrot} is true when the player has the carrot). We discuss how event detection occurs in \autoref{sec:approach}, and give an example of how progression works in \autoref{sec:app_progression}.
% Note that if at some later point the agent drops the carrot, the LTL formula will not be updated and regress to its previous form. This is because in the sequence\commentmt{confirm this and be sure of whether we want to inclde this}

% In order for progression to occur, it's necessary to have \textit{event detectors} that can detect when propositions (like \ltlpredicate{player-has-carrot}) are true as the agent acts during an episode. We discuss how event detection occurs in \autoref{sec:approach}.
%the subsequent sections.
\section{Following Instructions with GATA}
\label{sec:gata_follow}
In order to evaluate the effectiveness of state-of-the-art text-based game agents at following instructions, we conducted experiments on the Cooking domain using the state-of-the-art model-free RL agent for TextWorld, GATA \citep{adhikari2020learning}.
GATA uses a transformer variant of the popular LSTM-DQN \citep{narasimhan2015language} combined with a dynamic belief graph that is updated during game-play. The aim is to use this belief graph as long-term memory to improve action selection by modelling the underlying game dynamics \citep{adhikari2020learning}. Formally, given the POMDP, GATA attempts to learn some optimal policy $\pi^*(a | o, g)$ where $g$ is the belief graph.

% We found that GATA was often unable to complete tasks, and indeed its performance on tasks was largely unaffected by whether or not it was even given instructions to follow, as will be described below. The experimental setup is described in \autoref{sec:app_stripped}, and we describe the GATA model in greater detail in \autoref{sec:app_model}.

% GATA uses a transformer variant of the popular LSTM-DQN \cite{narasimhan2015language} combined with a dynamic belief graph that is updated during game-play. The aim is to use this belief graph to improve action selection by modelling the underlying game dynamics \citep{adhikari2020learning}. Formally, given the POMDP, GATA attempts to learn some optimal policy $\pi^*(a | s, g)$ where $g$ is the belief graph.
% GATA is among the popular trend \citep{ammanabrolu2020graph, zelinka2019building, ammanabrolu2018playing} of using memory to augment an agent's state space. 

While GATA's belief graph can capture goal relations (e.g. apple-needs-cut), it turns out that agents trained to condition on observations and the GATA belief graph alone largely ignore in-game instructions. We tested a GATA agent on levels 1 and 2 in the Cooking domain, after training on either the 20-game or 100-game training set, and found that in none of those settings was the cookbook examined more than 15\% of the time (3/20 testing games). In short, \emph{the GATA agent usually doesn't observe what the recipe is for the current game}, meaning it has no way of knowing what the actual goal of the game is (except -- eventually -- from the rewards it gets and when the episode ends).

We further investigate how GATA agents fail to follow instructions by training these agents using modified game observations that have their instructions stripped (specifically, instructions directing the agent to examine the cookbook, the recipe text within the cookbook, and instructions to grab a knife if attempting to cut an ingredient without first holding the knife were removed from observations). This has two effects: (1) the agent no longer receives text-based instructions about what the goal is or what it should do; and (2) GATA's belief state will no longer capture goal relations like `needs'. The results of this experiment are in \autoref{fig:stripped}, and demonstrate how GATA's performance remains largely unchanged. This suggests that GATA is (here at least) (a) \emph{not exploiting text-based instructions that would lead it to success} and (b) \emph{even not exploiting the goal-related relations in its own belief state}. 

% \added{How can an agent do as well as GATA without reading the recipe? There may be patterns in the data that can be exploited, but we should note that the Cooking domain is relatively forgiving. The agent is not penalized for collecting or preparing irrelevant ingredients (except that doing so delays getting rewards, and rewards are temporally discounted). It is still the case that if the agent prepares an ingredient needed by the recipe incorrectly (e.g., by dicing when the recipe called for slicing), then the episode will end immediately. However, if the recipe is simple enough, an agent could succeed at a non-trivial rate by collecting random ingredients and guessing how to prepare them, though that strategy would have a significant performance drop-off with even a slight increase in recipe size.}

The results in \autoref{fig:stripped} also show a drop in %the performance of
GATA's performance when moving from level $1$ to level $2$ in the Cooking domain, where the games' complexity is increased by just one added ingredient preparation step in the recipes (see \autoref{tab:levels} for more details on the %different
levels). GATA has difficulty in fully completing tasks on level 2 games, where its success rate is roughly half that of its achieved normalized game points (only the latter metric was used by \cite{adhikari2020learning}). 

Given these insights, we wish to further study and address instruction following in TBGs. In the next section, we propose using LTL and demonstrate how existing work can be easily augmented.

% Scott: 
\section{An Approach to Following Instructions}
\label{sec:approach}

We now investigate a mechanism for both studying and advancing the ability of an RL agent to follow instructions.  We do so by translating instructions to an internal structured representation of language in the form of LTL, a formal language that is increasingly being used for reward specification in RL agents~\citep{vaezipoor2021ltl2action, leon2020systematic, kuo2020encoding, camacho-etal-ijcai19,toro2018teaching}. We describe how to augment the GATA architecture with these LTL instructions and how to monitor progress towards their completion.

% PREVIOUS DISCUSSION, FOR REFERENCE
%build on the To date, the state-of-the-art (SOTA) model-free RL agent for TextWorld is the Graph Aided Transformer Agent (GATA) \citep{adhikari2020learning}. GATA uses a transformer variant of the popular LSTM-DQN \cite{narasimhan2015language} combined with a dynamic belief graph that is inferred and updated during planning. The aim is to use the dynamic belief graph to improve action selection by modelling the underlying game dynamics \citep{adhikari2020learning}. Formally, given the POMDP $<S, T, A, O, \Omega, R, \gamma>$, GATA attempts to learn some optimal policy $\pi^*(a | s, g)$ where $g$ is the belief graph of the agent.

\subsection{Generating and Representing LTL Instructions for TextWorld}
\label{sec:ltl_gen}
We use three types of instructions for the Cooking domain. The first instruction identifies the need to examine the cookbook: This instruction is defined as
\(
    \varphi: \ltlnextTXT \ltlpredicate{cookbook-is-examined}.
\)
This instruction simply states that the agent should examine the cookbook (i.e. $\ltlpredicate{cookbook-is-examined} = \true$) in the next step of the game. The second instruction is the actual recipe that gets elicited from the cookbook. We format this instruction to be \emph{order-invariant} and \emph{incomplete}. Order-invariance allows the agent to complete the instructions in any order, but is still constrained by any ordering that the TextWorld engine may enforce. ``Incomplete'' simply refers to the fact that not every single action required to complete the recipe is encoded (i.e. grabbing a knife before slicing a carrot, opening the fridge). The agent must still learn to do these things to accomplish its tasks, but is not directly instructed to. Assuming the recipe requires that predicates $p_1, p_2, \hdots p_n$ be true, the cookbook instructions are modelled as
\(
    \varphi: (\ltleventuallyTXT p_1) \wedge (\ltleventuallyTXT p_2) \wedge \hdots (\ltleventuallyTXT p_n).
\)

For example, in the Cooking Domain, this instruction might be the conjunction
\[
    \varphi: (\ltleventuallyTXT \ltlpredicate{\scriptsize{apple-in-player}}) \wedge 
             (\ltleventuallyTXT \ltlpredicate{\scriptsize{meal-in-player}}) \wedge 
             (\ltleventuallyTXT \ltlpredicate{\scriptsize{meal-is-consumed}}).
\]
The third and final type of instruction identifies the need to navigate to the kitchen. This instruction is defined as
\(
    \varphi: \ltleventuallyTXT \ltlpredicate{player-at-kitchen}.
\)
This instruction will come prior to the first two described above, but is only used in games with navigation (see \autoref{tab:levels}). 
% \commenttk{How does the player know to switch to the cookbook instruction after reaching the kitchen?}

We build a simple LTL translator that generates these instructions from the textual observations, similar to the goal generator used in \cite{liu2022learning}. TextWorld's observations are easily parsed to extract the goal information already contained within them, which we then formalize and keep track of using LTL. 
%We specifically only ever generate three instructions: (1) the need to check the cookbook, which is generated from the initial game observation; (2) the recipe instruction, which is generated from the observation following an agent examining the cookbook; and (3) an instruction to find the kitchen, which is also generated from the initial game observation and only applies on certain game level. 
We provide examples of these observations and more details in \autoref{sec:app_ltl_generation}. Note that these observations are only used to generate the instruction itself, and subsequently LTL progression is used with the GATA belief state as our event detector to monitor completion of instruction steps and to update instructions that remain to be addressed. 
% \commenttk{Maybe we should say something about ``event detectors'' here.}

%\added{
One possible criticism with such an LTL translator is its reliance on domain knowledge. While not the main focus of this paper, a complementary research problem that has begun to be explored is to \emph{automatically} translate natural language instructions to LTL \citep[e.g.,][]{scheutz2009, finucane2010ltlmop, wang2020learning}. Traditionally, such approaches have required large corpora of training data or hard-coded rules, and were restricted to a specific domain. However, pretrained large language models such as GPT-3 introduce the potential for a general natural-language-to-LTL translation scheme with minimal domain-specific adaptation \citep{hahn2022formal, huang2022language, brohan2022can}. We explore this prospect by applying GPT-3 to TextWorld in \autoref{sec:nl2ltl}. 
% We consider this beyond the scope of our work, \commenttk{If it's beyond the scope of our work, why did we run those extra experiments?} but we further discuss this prospect in \revisit{Section~\ref{sec:experiments}}\alt{\autoref{sec:nl2ltl}}.
%}

Finally, we note that in this work, GATA provides the domain-dependent vocabulary for describing properties of state (e.g. $\ltlpredicate{carrot-is-chopped}$) while our LTL augmentation provides the \textit{domain-independent} temporal modalities (i.e., $\ltlnextTXT, \ltleventuallyTXT$, etc.) and the logical connectives for composing those properties of state into the instructions we use. In this way, our technique is very generalizable, limited only by the recognizable properties of state (which in our case are provided by GATA) and instructions that can be extracted in game-play. 
% \commenttk{well, maybe also by where we can find instructions}, .

\subsection{LTL Augmented Rewards and Episode Termination}
\label{sec:reward_episode_termination}

We can also reward our agent for completing instructions, which we model as reward $\RLTL(s,a,\varphi)$. 
% \commenttk{Should we just call that $R'$? A reviewer was confused by what $\Phi$ stood for. Or maybe $R_\text{LTL}$?} \commentmt{i thought of that, i changed it for now, I think its an easy change that could alleviate some issues.} \commentsm{Mat, please make sure it's enforced consistently in the paper. You could use a macro and for now make it red as well as saying $\RLTL$ so that we can see the change, but then change the macro to remove the red.}\commenttk{Also don't forget the appendix}
For some labelling function $L : S \times A \rightarrow 2^{\mathcal{P}}$ that assigns truth values to the propositions in $\mathcal{P}$, 
\[
    \RLTL(s, a, \varphi)  =R(s, a) + 
    \begin{cases}
    1 ~~ \text{if}~ \mprog(L(s,a), \varphi) = \true\\
    -1 ~~ \text{if}~ \mprog(L(s,a), \varphi) = \false\\
    0 ~~ \text{otherwise}
    \end{cases}
\]
In other words, a bonus reward is given for every LTL instruction the agent satisfies and a penalty is given if the agent fails to complete an instruction. We perform an ablative study on the effect of this reward in \autoref{sec:exp_reward_term}. We henceforth refer to this modified reward function as the \textit{LTL reward}. The maximum bonus reward an agent receives is either $2$ if there is no navigation task, or $3$.

Further, because we wish to \textit{satisfy} instructions, 
we can also use the instructions to modify episode termination. That is, if our LTL instruction is %said to be 
violated, we 
have arrived in a terminal state, even if TextWorld has not indicated so. We perform an ablative study on the effect of this \textit{LTL-based termination} in \autoref{sec:exp_reward_term}.

\subsection{LTL-GATA Model Architecture}
\label{sec:model_architecture}
\begin{figure}[t!]
    \centering
    \includegraphics[width=0.9\textwidth]{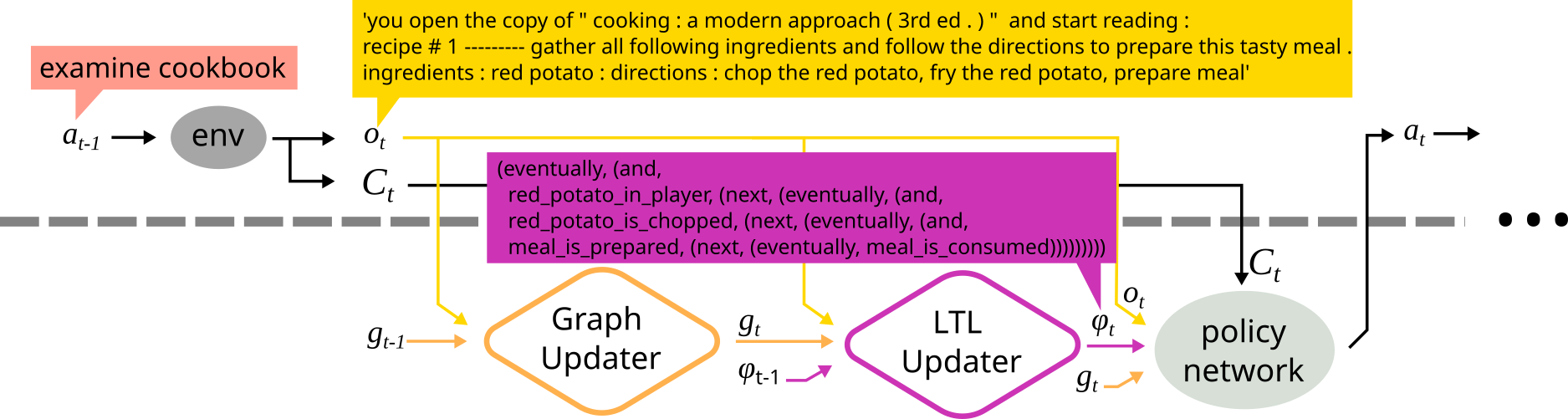}
    \caption{An example of a single step in an episode of TextWorld. The game environment returns an observation $o_t$ and action candidate set $C_t$ in response to action $a_{t-1}$. In turn, the agent's graph updater (GATA) updates its belief graph $g_t$ in response to both $o_t$ and $g_{t-1}$. Next, $g_t$ and $o_t$ update the LTL instructions. $\varphi_t$ is generated from $o_t$ after the cookbook is examined and thereafter $\varphi_{t-1}$ is progressed to $\varphi_{t}$ at each time step.
    The policy network selects action $a_t$ from $C_t$ conditioned on $o_t$, $\varphi_t$, and $g_t$ and the cycle repeats.} 
    \label{fig:data_flow}
\end{figure}

We build a similar model to GATA's original architecture, augmented to include the LTL encoding of instructions and their progression according to observed system state.  We dub this model \agentname, which we describe in detail below. \autoref{fig:data_flow} depicts an episode step interaction of LTL-GATA with TextWorld and Figure \ref{fig:model} depicts the model itself. Additional details can be found in \autoref{sec:app_model}.

\textbf{Graph Updater:} We use the original GATA-GTP model~\citep{adhikari2020learning}, which generates a discrete belief graph as a list of triplets of the form  $(\textit{object},\textit{relationship},\textit{object})$ 
% of the knowledge state of the world, e.g., $(\textit{red potato},\textit{is},\textit{chopped})$
. It is composed of two sub-components: (a) the belief state updater, which generates $g_t$ from observation $o_t$ and the graph $g_{t-1}$; and (b) the graph encoder, which encodes the current graph into a vector as $\texttt{GE}(g_t) = g'_t \in \mathbb{R}^D$ for some latent dimension $D$. The graph encoder is a relational graph convolutional network (R-GCN) \citep{schlichtkrull2018modeling} using basis regularization \citep{schlichtkrull2018modeling} and highway connections \citep{srivastava2015highway}. We refer the readers to \cite{adhikari2020learning} for more details.

\textbf{LTL Updater:}  
The LTL updater generates and progresses LTL instructions. % $\varphi_t$.  
LTL instructions defining the need to arrive at the kitchen and examine the cookbook are generated from the initial observation $o_0$. The subsequent instruction defining the recipe is generated from game observation $o_t$, as described in \autoref{sec:ltl_gen}, when the action $\textit{examine cookbook}$ is executed at time $t$. For the truth assignments (i.e. the labelling function $L$), we leverage GATA's highly accurate belief state from the graph updater.
We use the Spot engine \citep{spotsoftware} to perform the progression.

\textbf{Text Encoders:} For encoding the action choices $C_t$, observations $o_t$, as well as encoding the LTL instructions $\varphi_t$, we use a simplified version of the Transformer architecture presented by \cite{vaswani2017attention}. This is the same architecture used by \cite{adhikari2020learning}. For LTL instructions, we encode them directly as a string. For example, the LTL formula $\varphi: (\ltleventuallyTXT p_1) \wedge (\ltleventuallyTXT p_2)$ where $p_1=\ltlpredicate{pepper-in-player}$ and $p_2=\ltlpredicate{pepper-is-cut}$, has the string representation 
\begin{align*}
    \text{str}(\varphi): ``\text{eventually player\_has\_pepper and eventually pepper\_is\_cut}``
\end{align*}
We format each predicate as a single token, and we show in \autoref{sec:app_exp_pred} that our method is robust to predicate format. 
% The word embedding of each token is the average of the word embeddings for each underscore-separated token in the predicate.
For some input string $v \in \mathbb{R}^{\ell}$ of length $\ell$, the text encoder outputs a single vector $\texttt{TE}(v) = v' \in \mathbb{R}^D$ of dimension $D$, which is the same latent dimension as the graph encoder.

\textbf{Action Selector:}  The action selector is a 2-layer multi-layer perceptron (MLP). The encoded state vectors $\texttt{TE}(o_t) = o_t' \in \mathbb{R}^D$, $\texttt{TE}(\varphi_t) = \varphi_t' \in \mathbb{R}^D$, and $\texttt{GE}(g_t) = g_t'\in \mathbb{R}^D$ are concatenated to form the agent's final state representation $z_t = [o_t';\varphi_t';g_t'] \in \mathbb{R}^{3D}$. In contrast to \cite{adhikari2020learning}, we concatenate features rather than use the bi-directional attention-based aggregator. This simplified the model's complexity and worked just as well experimentally. This vector is then repeated $n_c$ times and concatenated with the encoded actino choices $C_t'\in \mathbb{R}^{n_c\times D}$ where $n_c$ is the number of action choices. 
% This concatenation works by repeating the state's vector representation $n_c$ times and then concatenating. 
This input matrix is fed to the MLP which returns the a vector of Q-values for each action $q_c \in \mathbb{R}^{n_c}$. %, which represents the Q-values for each action choice.
    % \begin{figure}[!t]
    % \begin{minipage}{0.7\textwidth}
    %     \centering
    %     \includegraphics[height=2cm]{figures/model/policy_model.pdf}
    %     \caption{\agentname's policy model. The model chooses action $a_t \in C_t$ conditioned on the state $z_t = [o_t';\varphi_t';g_t']$. The action selector chooses $a_t$ based on the predicted Q-values.}
    %     \label{fig:model}
    %     \end{minipage}
    % \end{figure}
\begin{table}
    
    \begin{minipage}{0.61\textwidth}
        \centering
        % \vspace{2mm}
        \includegraphics[height=2cm]{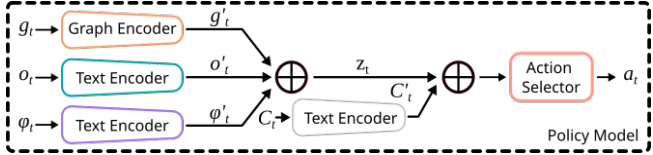}
        \vspace{0.5mm}
        \captionof{figure}{\agentname's policy model. The model chooses action $a_t \in C_t$ conditioned on the state $z_t = [o_t';\varphi_t';g_t']$. The action selector chooses $a_t$ based on the predicted Q-values.}
        \label{fig:model}
        \end{minipage}
        \hspace{2mm}
        \begin{minipage}{0.29\textwidth}
        \centering
        \vspace{-5mm}
    \caption{Cooking Levels}
    \small{
    \begin{tabular}{ccccc}%{||c|ccccc||}
        \toprule%\hline
         \rotatebox{90}{Level}&\rotatebox{90}{\parbox{1cm}{Recipe\\ Size}}&\rotatebox{90}{Rooms}&\rotatebox{90}{\parbox{1cm}{Max Score}}&\rotatebox{90}{\parbox{1cm}{Need \\ \{Grab, Cut, Cook\}}}\\
        \midrule% \hline
        $0$&$1$&$1$&$3$&\{\cmark,~\xmark,~\xmark\}\\
        $1$&$1$&$1$&$4$&\{\cmark,~\cmark,~\xmark\}\\
        $2$&$1$&$1$&$5$&\{\cmark,~\cmark,~\cmark\}\\
        $3$&$1$&$9$&$3$&\{\cmark,~\xmark,~\xmark\}\\
        % $4$&$3$&$1$&$11$&\cmark&\cmark\\
        \bottomrule%\hline
        % \hline
        % $5$&\multicolumn{5}{c}{Equal Mixture of Levels $\{0, 1, 2, 3, 4\}$}
    \end{tabular}
    \label{tab:levels}
    }
    \end{minipage}
    % \vspace{-6mm}
\end{table}

 \textbf{Training.} Formally, for belief state $g$ and LTL instruction $\varphi$, \agentname\ aims to learn an optimal policy $\pi^*(a|o, g, \varphi)$. To learn this optimal policy, we implement Double DQN (DDQN) \citep{van2016deep} with reward function and termination criteria as discussed in \autoref{sec:reward_episode_termination}. We use a prioritized experience replay buffer \citep{schaul2015prioritized}. Refer to \autoref{sec:app_training} for further details.

\section{Experiments}
\label{sec:experiments}

% \commenttk{We conduct experiments to answer the question of whether our approach can outperform baselines including GATA in the Cooking domain, in particular in terms of successfully completing the game and not just game points...}
% \commentsm{We also perform ablations to determine ...}

Our experimental assessment was designed both to understand how well GATA 
%, as an example of a SOTA TBG agent, 
was exploiting observational instructions, as discussed in \autoref{sec:gata_follow}, and to assess the instruction-following performance of our proposed approach relative to this state of the art (not only in terms of game points but also successful completion). We additionally strove to assess features of our approach (such as monitoring instruction progress) that contributed to its performance, as well as general challenges to text-based game playing that limited its performance (such as navigation).\footnote{Our code for the experiments can be found at \codeurl}

% \remove{We conduct experiments to determine whether our approach can outperform baselines, including GATA, in the Cooking domain, with a particular focus on success rate, increasing game complexity, and increasing dataset size. We also perform ablations to determine the importance of progression as well as whether GATA can take advantage of the cookbook.}

% take step back, broader - instruction following in tbgs

\subsection{Experimental Setup}

\textbf{Games.}
To have as fair a comparison with \cite{adhikari2020learning} as possible, we reused the sets of games they had generated. For the training games, they had created two sets: one set that contains 20 unique games per level and another that contains 100 unique games per level. Both the validation and testing sets have 20 unique games each per level. 
The levels we chose to use in our assessment are shown in \autoref{tab:levels}. Note that in our assessment we omit Levels 4 and 5. Level 4 is an augmentation to Level 3 that adds more ingredients; both GATA and LTL-GATA at this level suffer from the navigation issues we discuss later with respect to Level 3. As we wanted to focus on instruction following and not navigation, we omitted this level and chose to use Level 0 instead. Level 5 is simply a random combination of all levels, 
%and \revisit{its performance} is dominated by the weak navigation performance, 
so it is omitted for similar reasons.
%}

% Note that GATA also reported results for a Level 4 (which is the same as Level 3 but with more ingredients and more preparation tasks) and a Level 5 (which is a random combination of Levels 1 through 4). Level 4 was omitted because it poses a very challenging navigation problem that is beyond the scope of instruction following we wish to focus on in this work. We note that both GATA and our agents struggle in the level, as similarly report in \cite{adhikari2020learning}. Similarly, Level 5 is built using Level 4 and is omitted for similar reasons.

% \textbf{Hyper-parameters.} To remove any bias towards more finely tuned experimental configurations, we replicate all but two hyper-parameters from the original GATA work: (1) we use a batch size of 200 instead of 64 when training on the 100 game set, (2) for level 3, we use Boltzmann Action selection. These changes boosted performance for all models. See \autoref{sec:app_hp} for more details.
\textbf{Hyper-parameters.} We replicate all but three hyper-parameters from \cite{adhikari2020learning}: (1) we use a batch size of 200 instead of 64 when training on the 100 game set, (2) for level 3, we use Boltzmann Action selection, and (3) we use Adam \cite{kingma2014adam} with a learning rate of $0.0003$ instead of RAdam \cite{DBLP:conf/iclr/LiuJHCLG020} with a learning rate of $0.001$. These changes boosted performance for all models. See \autoref{sec:app_hp} for more details.

\textbf{Baselines.} We compare against (1) TDQN \citep{adhikari2020learning}, the transformer variant of the LSTM-DQN \citep{narasimhan2015language} model, (2) GATA\textsuperscript{C}, and (3) GATA\textsuperscript{D}. GATA\textsuperscript{C} is GATA's best performing model (GATA-COC) that uses a continue graph-updater pre-trained using contrastive observation classification. GATA\textsuperscript{D} is a similarly performant model (GATA-GTP) that uses a discrete graph-updater pre-trained with ground-truth graphs from the FTWP dataset. Finally, we note that we found a few issues with GATA's original code\footnote{https://github.com/xingdi-eric-yuan/GATA-public, released under the open-source MIT License.} and have since fixed them (see \autoref{sec:app_fixed_gata}). For comparison, we include the original \textit{paper} GATA models, labelled as $\text{GATA}^{\text{C}}_{\text{P}}$ and $\text{GATA}^{\text{D}}_{\text{P}}$.

\textbf{Measuring performance.} We measure performance using two metrics: normalized accumulated game points and game success rate. We report averaged results over $3$ seeds for each experiment. Previous works only compared using the normalized accumulated game points; however, this may sometimes be misleading --- an agent could get $3/4 = 0.75$ points on all games but never actually succeed on any. 
% Obvious of previous comment?  -Scott
% agreed.
%A normalized score of $0.75$ may look promising but doesn't identify how the agent is in fact always failing at fully completing the task. 
In contrast, measuring the success rate alongside the normalized game points allows for a more complete analysis of the agent's ability to play and complete these games.

\subsection{LTL-GATA Compared to Baselines}

\begin{figure}[t]
    \centering
    \includegraphics[height=3.5cm]{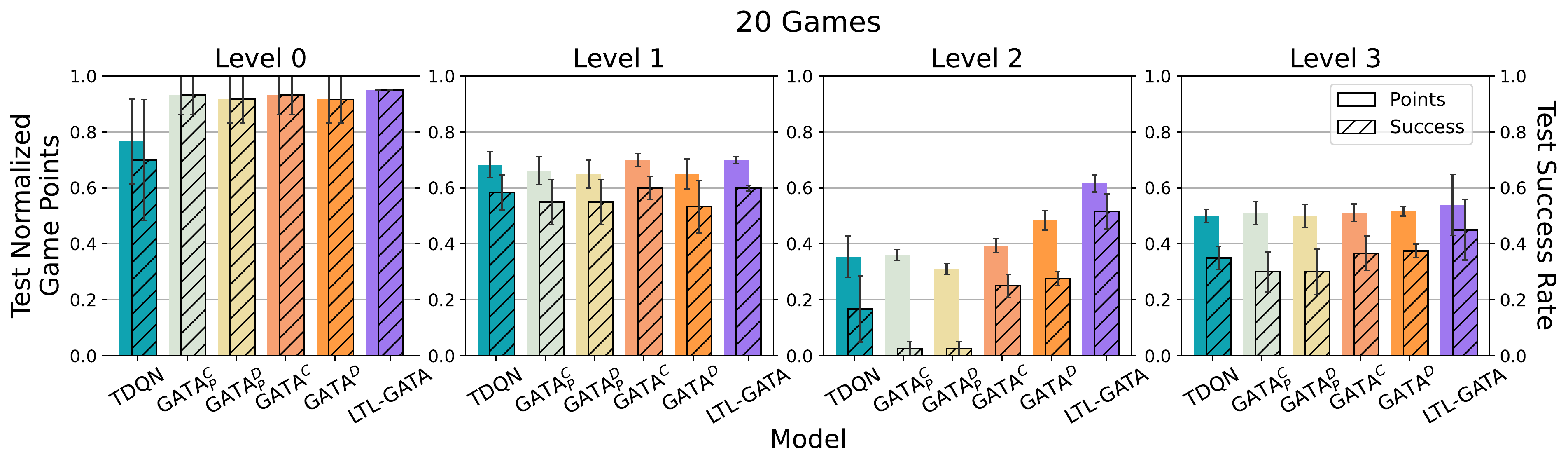}\\
    \includegraphics[height=3.5cm]{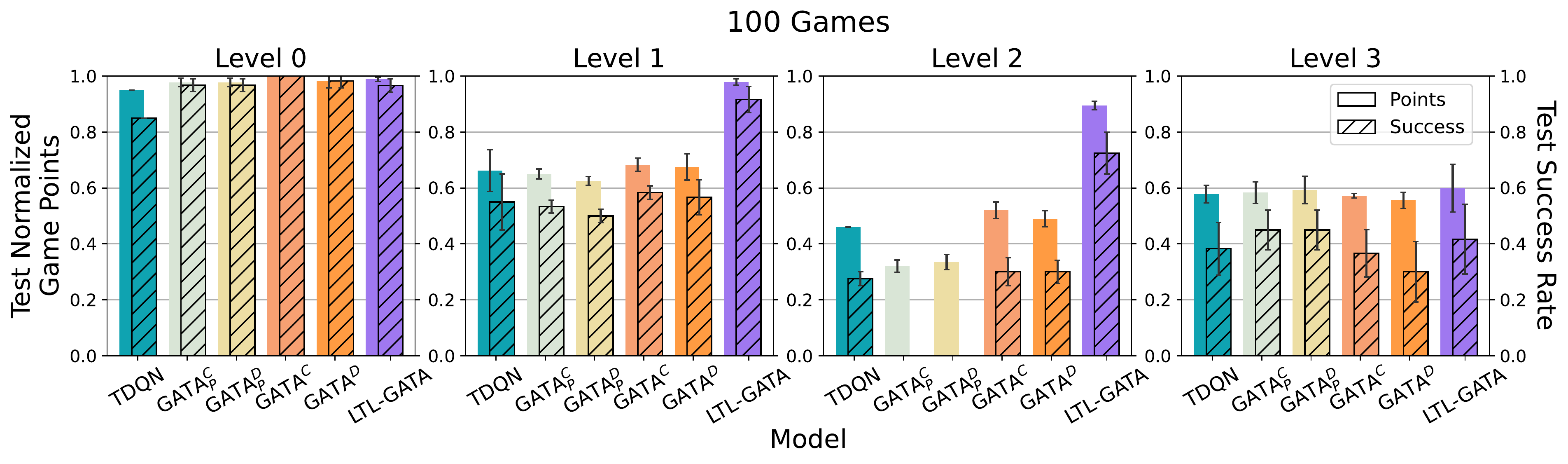}
    \caption{Testing scores across various levels and on both the 20 (top) and 100 (bottom) game training sets. We select the top-performing models (per seed) on the validation set during training and apply those models on the test set and report the average scores.} %. \agentname\ outperforms previous state-of-the-art, with particularly large gains on the 100 game training set.}
    \label{fig:level_perf}
\end{figure}
\begin{figure}
    \centering
    \begin{subfigure}{0.75\textwidth}
    \includegraphics[height=2.8cm]{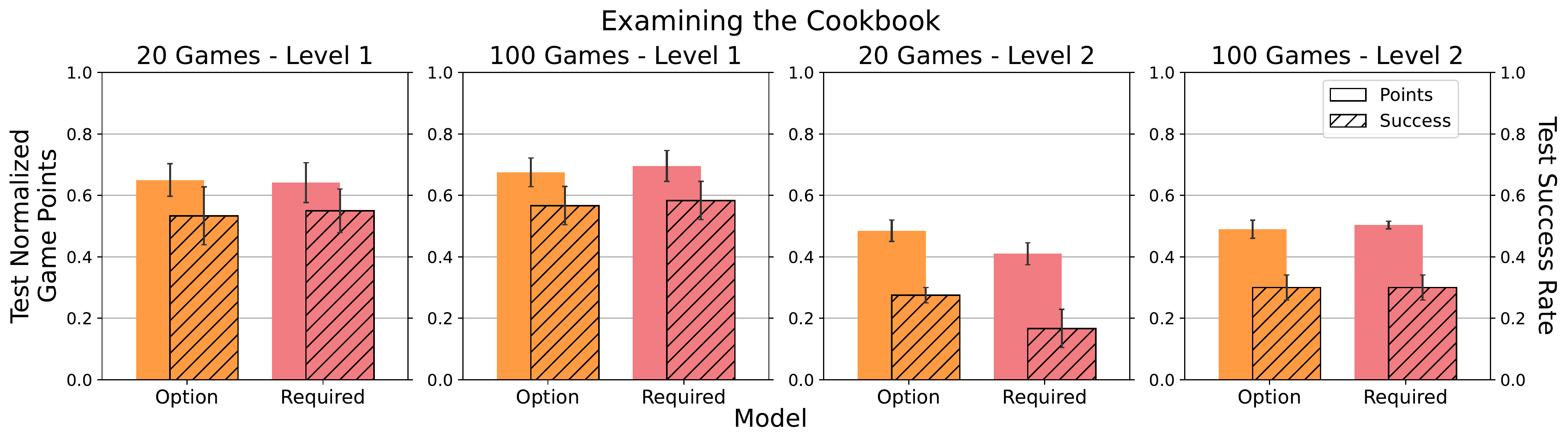}
    \caption{Forcing GATA to Examine the Cookbook}
    \label{fig:cookbook_gata}
    \end{subfigure}
    \begin{subfigure}{0.24\textwidth}
    \includegraphics[height=2.8cm]{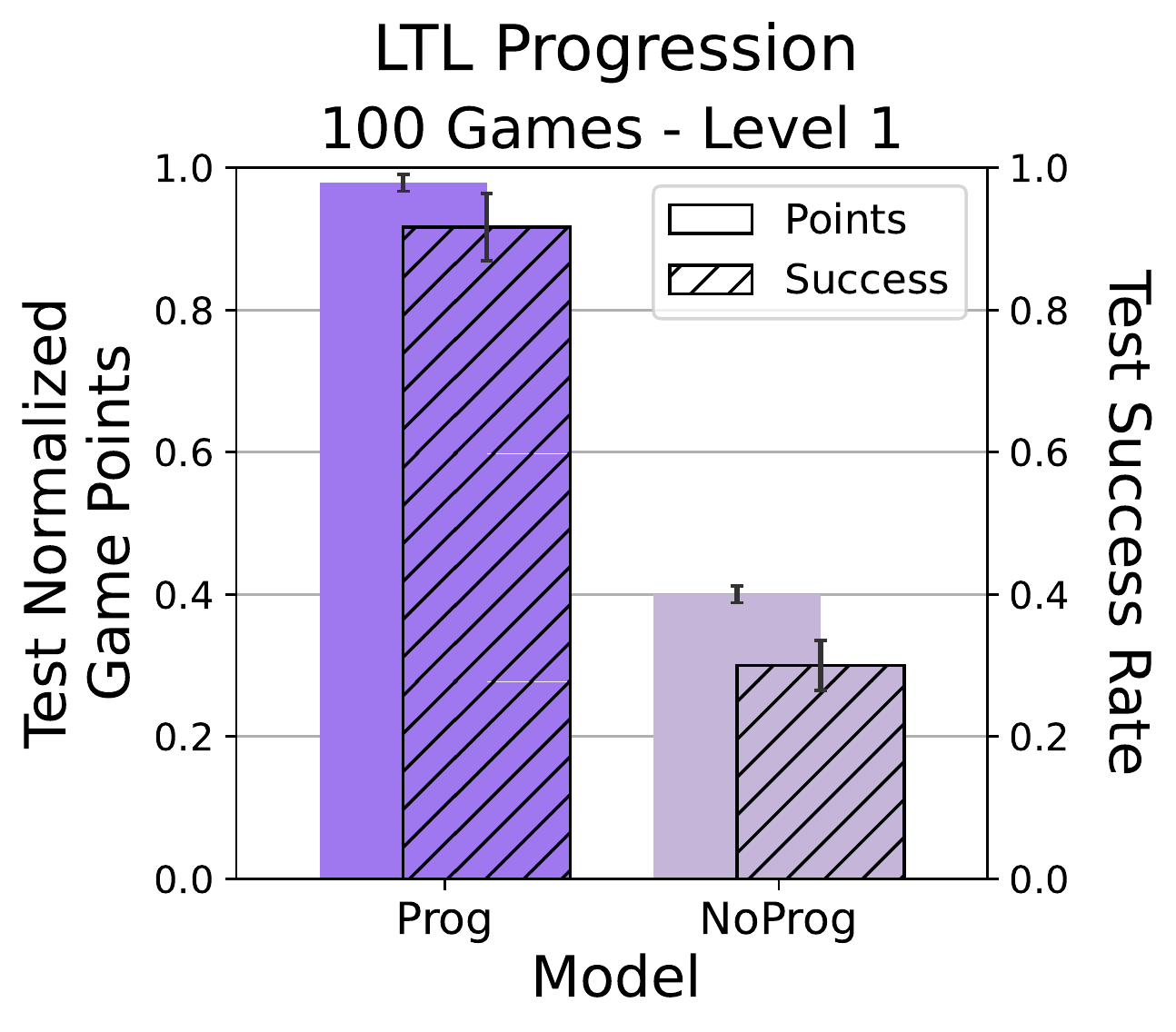}
    \caption{Progression Ablation}
    \label{fig:prog}
    \end{subfigure}
    \caption{(a) A comparison of GATA\textsuperscript{D} performance when given the \emph{Option} to examine the cookbook vs. when it is \emph{Required} to examine the cookbook. (b) A comparison of LTL-GATA with (\emph{Prog}) and without (\emph{NoProg}) using LTL progression.}
    \label{fig:cook_prog}
    % \vspace{-5mm}
\end{figure}
\textbf{Consistently high performance with 20 training games}. We see from \autoref{fig:level_perf} that \agentname\ exhibits consistently high performance across levels as compared to the baselines when trained on the 20 games set. In particular, \agentname\ maintains its performance on level 2, where the game's slight increase in complexity causes large performance drop-offs in other methods. Our agent can easily complete the added task and maintain similar performance to the previous level 1.

\textbf{Large performance gains with 100 training games}. We see from \autoref{fig:level_perf} that \agentname\ gains considerable performance when trained on 100 games. With the added games, our agent is exposed to more predicates and can now generalize better to the testing set. Future work may look at how to achieve this kind of generalization without having to expose our agent to more predicates.

\textbf{Success rate and normalized game points.} Looking at the performance of GATA on level 2, it becomes apparent why measuring the success is important. Although it achieves almost $0.4$ normalized points, the actual success rate is near $0$ for original GATA models, and $\sim60\%$ of the normalized points for the fixed models average across both training sets. In contrast, \agentname\ exhibits high normalized points and success rate, where the average success rate across both training sets is $\sim82\%$ of the normalized points.

% \commentsm{**IMPORTANT** I just added this here. Please rephrase so that it fits better (as you see fit) and consider whether this is a good placement for this remark.}

% \added{
% \textbf{Generalization.} Interestingly, the structured relational representation afforded by the GATA belief graph supports generalization. Indeed, our experiments demonstrated generalization to unseen recipes in the Cooking domain, including novel combinations of objects and relations/verbs. Successful application of LTL in partially observable environments with imperfect propositional state estimation is an important milestone for this area of RL research.
% }

% \commentsm{I'm good with the red text.} \commentsm{The use of LTL for navigation -- a huge barrier -- could be mentioned in the conclusions as a topic of future work, if we have space}
\textbf{Competitive performance on level 3.} Level 3 introduces the added challenge of navigation. \agentname\ outperforms GATA in this level as well, but not to the degree of previous levels. Inspecting testing trajectories, it becomes evident that both \agentname\ and GATA methods struggle with navigation in this level, and have difficulties even navigating to the kitchen in the first place.
Exploring at test time to find items and rooms in an unknown environment is a major challenge built into text-based games. Hypothetically, LTL could contribute to addressing this challenge. LTL could be used to dictate strategy and/or to simply track such exploration (e.g., for remembering which rooms have been previously visited). LTL might also be used to encode \textit{learned} navigation instructions (e.g. ``find the blue door, go through it, then turn right''). We do not pursue this vector of research here, but it is an interesting direction for future work.

\subsection{Does LTL Progression Matter?}
% \begin{wrapfigure}[8]{R}{0.3\textwidth}
%     \centering
%     \vspace{-20mm}
%     \includegraphics[height=3.5cm]{figures/test-bar-plots/ltl_progression_test_bars.pdf}
%     \caption{A comparison of LTL-GATA performance with (\emph{Prog}) and without (\emph{NoProg}) using LTL progression.}
%     % \vspace{-12mm}
%     \label{fig:prog_ablation}
% \end{wrapfigure}

We show in \autoref{fig:prog} that the use of progression is critical to performance, where \agentname\ without progression incurs a large performance drop-off, dropping below the performance of the baselines as well. Without progression, the LTL instruction will not reflect the changes incurred by the agent's actions. This appears to confuse the agent considerably, demonstrated by its performance drop-off.

% \subsubsection{Ablation: LTL with Ground Truth Graphs vs. GATA}
% -had to rerun
% \commentsm{Please replace "Consume" with another word?}
\subsection{Forcing GATA to Examine the Cookbook}

Because LTL-GATA is always tasked with examining the cookbook, we question whether a similar tasking for GATA improves performance. We experiment with GATA\textsuperscript{D} by forcing the agent to examine the cookbook on the first step of the episode. Forcing GATA to examine the cookbook will elicit goal relations like $\ltlpredicate{(apple,needs,cut)}$ in the belief state. We show however in \autoref{fig:cookbook_gata} that GATA does not improve when being given the cookbook. This shows that GATA cannot make use of the information elicited from the cookbook, continuing to ignore important instructions. Even with the presence of goal relations in its belief state, GATA fails to properly attend to this information. This highlights the benefits of a formalized representation of instructions used by \agentname.

\subsection{On Automatic Translation: Natural Language Instructions to LTL}
\label{sec:nl2ltl}

% \commentsm{IMPORTANT -- YIKES - are paragraph2 and paragraph3 the old and the new (aka "alt") version of the same thing??}
% \commentmt{they were, i removed it p2, not sure how it got put back in I swear I didn't see it before}

% \commentsm{I like this section (that follows) and vote for leaving it in, Scott's comments notwithstanding.  Please make it all black once you've vetted it.}

While LTL-GATA relies on a handcrafted LTL translator to provide initial instructions from text observations, we investigate the potential of automating this step using pretrained large language models. This is not a central focus of the paper. Rather, we include this exploration as a proof of concept that the use of LTL is not a barrier to broad deployment of the work presented here. To this end, we evaluate whether GPT-3 \citep{brown2020language} can few-shot learn to translate TextWorld observations to LTL, given only \emph{six} examples and without additional training. 

% Dataset details
% Prompt examples
% Example of DaVinci generalization
% Example of Ada failure

% We experiment with two models of GPT-3: \emph{Ada} (the fastest model) and \emph{Da Vinci} (the most powerful model). To perform a translation, each model completes a prompt containing six examples of the form "NL: \texttt{\{natural language observation\}}. LTL: \texttt{\{LTL formula\}}" followed by "NL: \texttt{\{observation to be translated\}}. LTL:". We consider a response that exactly matches the ground-truth LTL formula as \emph{absolutely correct}, a response that is otherwise correct except for parentheses and spaces as \emph{almost correct}, and all other responses as \emph{incorrect}. Further experimental details and examples can be found in the \autoref{sec:nl2ltldetails}.%\commenttk{Which section of it?}

% \commentsm{I vote to use this version. If Andrew/Mat agree, let's make it black and comment out what it replaces.}
We experiment with two models of GPT-3 from OpenAI: \emph{Ada} (the fastest model) and \emph{Da Vinci} (the most powerful model). 
We perform few-shot translation by constructing prompts that contain six example translations, followed by the natural language observation to translate (the test case). 
% \commentmt{does this sentence feel repetitive with the last sentence from the paragraph above?} \commenttk{It shows that the six examples are part of the prompt, and not e.g. used to finetune GPT's parameters.}
The examples remain fixed for all test cases, and follow the form "NL: <\texttt{natural language observation}>. LTL: <\texttt{ltl-formulas}>". Our test case follows the form "NL: <\texttt{natural language observation}>. LTL:", where the model must complete the prompt, thereby performing a translation.
We consider a response that exactly matches the ground-truth LTL formula as \emph{absolutely correct}, a response that is otherwise correct except for parentheses and spaces as \emph{almost correct}, and all other responses as \emph{incorrect}. Further 
% experimental % saves that one line, brings conclusion up % good!
details and examples 
can be found 
%are % could save one line here
in \autoref{sec:nl2ltldetails}.

% \commentmt{You say 234 here, but "six" examples before. I'm just a little confused as to what specifically was fed to the models then.}
%%% Scott: I would say Ada got 0% correct to reinforce the drawbacks of weaker language models... switching to incorrect and saying 100% throws me off when I read the following paragraph.
% \commentss{Overall, this section seems quite compressed... would it be better just to leave it to the Appendix and cite the main results here... that GPT-3 Da Vinci can get 93.2 absolutely correct?  Up to you all.}
Out of 234 test cases, \emph{Da Vinci}
% managed to 
translated 93.2\% \emph{absolutely correctly} and another 5.6\% \emph{almost correctly}, with only 1.3\% of examples incorrect. \emph{Da Vinci} displayed an impressive ability to generalize to unseen adjectives (e.g. \texttt{is\_grilled}), nouns (e.g. \texttt{carrot}), and compositions of formula. Unfortunately, the weaker model, \emph{Ada}, translated all 100\% of examples incorrectly. We found that \emph{Ada} commonly hallucinated new nonsensical words and predicates such as \texttt{ingredient\_is\_salt\_is\_diced} or \texttt{banana\_pork\_chop\_in\_player}, leading to erroneous translations.

%  \commentmt{My justification for one level is that only one example is needed to show how removing both incentives/methods decayings performance considerably. Multiple examples can further show decaying performance but the evidence was clear from the first level, in my mind. Perhaps I'm wrong?. If we feel that isn't a strong enough argument, we can defer to the supplementary material ("we show further examples of this in the appendix...") and I can get those figures in for next week.}

% In general, that the reward has positive benefits, but removing that reward can in some instances actually improve performance (i.e. Level 2 on 20 games). We can conclude that the bonus reward for realizing LTL instructions isn't as important as the inclusion of the instructions themselves.

% Similar to the reward, we can see from \autoref{fig:termination_ablation} that episode termination has positive benefits on the performance of \agentname\, but does significantly degrade its performance. \agentname\ without LTL-augmented episode termination performs marginally worse than \agentname\ with termination on 20 games, however remains equally performant on 100 games.
\section{Related Work}
\label{sec:related_works}

\textbf{Text-based games.} In this work we equip a text-based deep RL agent with formalized LTL instructions, building on previous works that employed belief graphs for solving text-based games. \cite{adhikari2020learning} focused on supervised (i.e. translation) and self-supervised learned mechanisms to construct such belief graphs, whereas \cite{ammanabrolu2020graph, yin2019learn, ammanabrolu2018playing} employed rule-based methods. At a larger scope, there is a host of other works on playing text-based games using deep reinforcement learning  \citep{hausknecht2020interactive, zahavy2018learn, jain2020algorithmic, yin2019comprehensible}. \cite{yuan2018counting} used count-based memory to shape the reward to improve in exploration and generalization in a simple domain. 
% \remove{\cite{narasimhan2015language, he2015deep}} 
\cite{narasimhan2015language} and \cite{he2015deep} proposed variations of an LSTM-based model, which the TDQN model used in this work is built from. In just published work, \cite{liu2022learning} took a model-based approach, focusing on object-oriented dynamics. However, these works do not address the role and representation of instructions that defines our work. \cite{DBLP:conf/emnlp/KimuraOCKWATMG21} does employ a neuro-symbolic RL method using Logical Neural Networks. However, it does not focus on instructions, operates over all logical facts of the environment, and is applied to a simpler domain.

\textbf{Instruction following and Linear Temporal Logic.} \cite{vaezipoor2021ltl2action} trained an RL agent to follow various LTL instructions in both discrete and continuous action-space visual environments. They used R-GCNs to learn representations of the LTL instructions and also employed LTL progression. Their model showed good generalization performance on similar and much larger unseen instructions than those observed during training 
%(from a set of up to $\sim 10^{39}$ unique tasks)
.
% A task-agnostic pretraining scheme was also used to reduce the overhead of learning LTL semantics, which showed improvement in sample-efficiency of the downstream task. 
However, in contrast to the work presented here, they relied on ground-truth event detectors  and operated in fully observable settings, while %where 
%\remove{we use a learned event detector (GATA)}\added{
we use GATA's learned belief graphs, in a partially observable setting, to evaluate the truth or falsity of propositions and to progress formulae. We further distinguish ourselves from this work by opting for training the LTL semantics end-to-end using a transformer rather than an R-GCN.
% \comment{\cite{vaezipoor2021ltl2action} employed LTL in a fully-observable, visual, multi-task environment using ground-truth event detectors. They further pre-trained their LTL encoder. We differ not only in environment but using a learned event detector (GATA) and learning to encode LTL instructions end-to-end.}{For pashootan to boast on LTL2action here and polish this}.
Works using LTL for reward specification \citep{leon2020systematic, kuo2020encoding, camacho-etal-ijcai19, toro2018teaching, DBLP:journals/corr/LittmanTFIWM17} or advice \citep{icarte2018advice} in RL agents exist, however they do not focus on text-based environments nor partially observable ones.
%Additionally,  our work differs in that
% \revisit{we model instructions dynamically on the environment's observations and use a learned event detector (GATA) rather than ground truth.}\commentmt{repeated, see middle of this paragraph}

%\commentsm{I am adding some of this to the discussion of LTL2Action as well to make clear the novelty wrt lTL2Action specifically.}
%\added{
%\textbf{Relating to novelty (maybe move to intro)} To our knowledge, the architecture proposed here is the first to propose formal language exploitation within a text-based game setting. \remove{Previous works \citep{vaezipoor2021ltl2action} are deployed in fully observed settings, and assume the existence of "event detectors" that establish ground-truth propositional values (e.g., $\ltlpredicate{carrot-is-chopped}$) at both training and testing time. This is in contrast to our work that uses learned belief graphs (through GATA) to establish these propositional values and progress instructions.} Interestingly, the structured relational representation afforded by the GATA belief graph supports generalization. Indeed, our experiments demonstrated generalization to unseen recipes in the Cooking domain, including novel combinations of objects and relations/verbs. Successful application of LTL in partially observable environments with imperfect propositional state estimation is an important milestone for this area of RL research.
%} \commenttk{There's some overlap between this and the previous paragraph.}
\section{Conclusion}
\label{sec:conclusion}

% Scott's notes from Tuesday discussion

% Conclusion: text-based games are an environment for studying an importance class of progress

% instruction following is key but has not been studied

% we show that existing systems are not following instructions

% we've proposed a way to internally represent instructions and track progress and provide this to an RL agent that led to significant improvement for TextWorld

% (instructions make no difference)

% (always want to use formal languages where semantics is well-defined)

% (we've inherited limitations of the architectures we build on)

% ========================================
% TextWorld acts as a sandbox learning environment for the larger class of problems dealing with a machine's ability to interact and act in dynamic systems with natural language. 
%In this work
%In this paper we investigated the task of instruction following by RL agents in text-based game environments. We conducted experiments that showed how current state-of-the-art model-free agents largely ignore observational instructions, and do not typically complete prescribed tasks. We proposed an approach to instruction following 

We studied the ability of RL agents to follow instructions in text-based games using TextWorld. We conducted experiments to show how current state-of-the-art model-free agents largely fail to exploit instructions and do not typically complete prescribed tasks. 
%generally ignore instructions 
We then showed how LTL can be used to construct internal structured representations for state augmentation that result in large performance improvements and more reliable instruction following and task completion.
% We showed the benefits of using LTL and its progression operator and made progress towards building RL agents capable of solving these games. 
%
%
%
%We found that a barrier to game play study and performance in RL continues to be effective navigation.
%
Experiments showed that monitoring instruction progress was critical to these gains.
Our method inherits limitations in dealing with navigation and unseen games from prior work, but these concerns are somewhat orthogonal to our focus on instruction following.
%
%Our method inherits limitations in dealing with navigation and unseen games from prior work, but these concerns are somewhat orthogonal to our focus on instruction following in this work. 
%We additionally do not have 
%
% ADD BACK
% Additionally, a general approach for English-LTL translation remains a general open question.
%
% Scott: I was confused by this discussion and just removed it... no need to go overboard on limitations.  I don't think the continuous issue is worth mentioning... also not entirely clear what was being discussed here "continuous belief" or "belief over continuous state"... also "belief" implies uncertainty to me!
%and our method cannot handle beliefs over continuous states.
%continuous belief states that have uncertainty. 
% \remove{These present important complementary directions for future work.}

%Finally, 
Furthermore, we can consider the broader impact of this work by relating to the critical need for good instruction following in safety-oriented domains such as autonomous transport or health care. 
%
%Furthermore, we 
We would like to suggest that works towards building better language agents should also emphasize the importance of \emph{completing} instructions. To illustrate, for an agent to help a person half-way across a street, or to start but not finish a medical operation, may be worse than for it to do nothing at all. To that end, we have proposed using (game) success rate as a metric for future work, and demonstrated how \agentname\ is very successful in the games it plays, relative to the state-of-the-art.
 Overall, we intend this paper to highlight the importance of studying instruction following
%We wish to bring importance to the need to study instruction following 
in environments like TextWorld that act as a proxies to the general class of problems dealing with language understanding and human-machine interaction.

Finally, in follow-on work we would like to explore more complex text-based games such as the Jericho environment \citep{hausknecht2020interactive}. These games involve a number of distinct challenges, including exploration, navigation, puzzle solving, language understanding, and instruction following. In this vein, we'd like to see whether LTL can be exploited to capture (learned) domain-specific strategic advice, or memory to tackle both navigation and exploration challenges. We'd like to further explore seamless ways to exploit the merits of natural language together with the benefits afforded by the compositional syntax and semantics of formal languages such as LTL. To this end, further advancing our explorations translating natural language to LTL is of interest and import, for this and a diversity of other applications in and outside RL.

\section*{Acknowledgements}

We thank the NeurIPS reviewers for their constructive feedback, and also the reviewers from the \emph{Wordplay: When Language Meets Games} workshop at NAACL 2022, where a preliminary version of this paper appeared \citep{TuliWordplay2022instruction}. %We were unable to address a number of them in the short time we had between notification and camera-ready, but they will be addressed in a future reporting of this work.
We gratefully acknowledge funding from the Natural Sciences and
Engineering Research Council of Canada (NSERC), the Canada CIFAR AI Chairs
Program, and Microsoft Research. Resources used in preparing this research
were provided, in part, by the Province of Ontario, the Government of
Canada through CIFAR, and companies sponsoring the Vector Institute for
Artificial Intelligence (\url{www.vectorinstitute.ai/partners}). Finally, we
thank the Schwartz Reisman Institute for Technology and Society for
providing a rich multi-disciplinary research environment.

% \newpage
% move to appendix
% \input{sections/broader-impact}

\newpage
\bibliography{citations}
\bibliographystyle{icml2021}

\newpage
\section*{Checklist}

% \commenttk{The section numbers in the checklist are hardcoded for some reason. We should make sure they still match the actual sections.} \commentmt{I went ahead and added the autorefs}

\begin{enumerate}

\item For all authors...
\begin{enumerate}
  \item Do the main claims made in the abstract and introduction accurately reflect the paper's contributions and scope?
    \answerYes{} 
  \item Did you describe the limitations of your work?
    \answerYes{} \textbf{Limitations are described in both \autoref{sec:experiments} and \autoref{sec:conclusion}.}
  \item Did you discuss any potential negative societal impacts of your work?
    \answerYes{} \textbf{A broader impact is considered in \autoref{sec:conclusion}, and further discussion can be found in \autoref{sec:broader}.}
  \item Have you read the ethics review guidelines and ensured that your paper conforms to them?
    \answerYes{}
\end{enumerate}

\item If you are including theoretical results...
\begin{enumerate}
  \item Did you state the full set of assumptions of all theoretical results?
    \answerNA{}
	\item Did you include complete proofs of all theoretical results?
    \answerNA{}
\end{enumerate}

\item If you ran experiments...
\begin{enumerate}
  \item Did you include the code, data, and instructions needed to reproduce the main experimental results (either in the supplemental material or as a URL)?
    \answerYes{} \textbf{All code can be found at \codeurl}
  \item Did you specify all the training details (e.g., data splits, hyperparameters, how they were chosen)?
    \answerYes{} \textbf{Hyper-parameters parameters are discussed briefly in \autoref{sec:experiments} and in full detail in \autoref{sec:imp_details} and \autoref{sec:app_hp}.}
	\item Did you report error bars (e.g., with respect to the random seed after running experiments multiple times)?
    \answerYes{} \textbf{Yes, error bars can be found in every results figure.}
	\item Did you include the total amount of compute and the type of resources used (e.g., type of GPUs, internal cluster, or cloud provider)?
    \answerYes{} \textbf{This information can be found in \autoref{sec:app_results}.}
\end{enumerate}

\item If you are using existing assets (e.g., code, data, models) or curating/releasing new assets...
\begin{enumerate}
  \item If your work uses existing assets, did you cite the creators?
    \answerYes{} \textbf{The footnotes in \autoref{sec:experiments} include links to the original assets, and in-text citations give credit throughout the paper.}
  \item Did you mention the license of the assets?
    \answerYes{} \textbf{The footnotes in \autoref{sec:experiments} include links to the original assets.}
  \item Did you include any new assets either in the supplemental material or as a URL?
    \answerYes{} \textbf{The only new asset in this work is our code, which we provide here: \codeurl.} 
    % \commenttk{Are we still going to put the code in the supplemental material, or only on github? Also, are we providing all the code?} \commentmt{only on github, and yes I believe so. Andrew gave me the GPT3 code which i will add to the repo.}
  \item Did you discuss whether and how consent was obtained from people whose data you're using/curating?
    \answerNA{}
  \item Did you discuss whether the data you are using/curating contains personally identifiable information or offensive content?
    \answerNA{}
\end{enumerate}

\item If you used crowdsourcing or conducted research with human subjects...
\begin{enumerate}
  \item Did you include the full text of instructions given to participants and screenshots, if applicable?
    \answerNA{}
  \item Did you describe any potential participant risks, with links to Institutional Review Board (IRB) approvals, if applicable?
    \answerNA{}
  \item Did you include the estimated hourly wage paid to participants and the total amount spent on participant compensation?
    \answerNA{}
\end{enumerate}

\end{enumerate}

\newpage
\appendix
\appendixpage
\startcontents[sections]
\printcontents[sections]{l}{1}{\setcounter{tocdepth}{2}}
% \section*{Learning to Follow Instructions in Text-Based Games: Supplemental Material}
% \begin{center}
%     \textbf{\LARGE{Learning to Follow Instructions in Text-Based Games}}
    
%     \vspace{-1.5mm}
%     \rule{0.25cm}{2pt}
    
%     \textbf{\LARGE{Supplemental Material}}
% \end{center}
% The contents of the supplemental material are as follows:
% \begin{itemize}
%     \item In \autoref{sec:app_rl} and \autoref{sec:app_pomdp}, we provide further background on Reinforcement Learning and Partially  Observed Reinforcement Learning, respectfully.
%     \item In \autoref{sec:app_tw}, we provide examples of observations in the TextWorld Cooking domain and highlight where the instructions come from and where the rewards come from.
%     \item In \autoref{sec:app_progression} we give an example of LTL progression.
%     \item In \autoref{sec:app_ltl_generation} we provide some examples of the LTL generation from instructions.
%     \item In \autoref{sec:app_model} we describe in greater detail the model architecture for LTL-GATA.
%     \item In \autoref{sec:app_exp} we describe in greater detail the experiments used in this work (e.g., hyper-parameters), provide some additional ablative studies, as well as provide the training curves for all experiments run.
%     \item In \autoref{sec:broader} we provide additional discussion on the broader impact of this work and potential negative societal impacts.
% \end{itemize}
% We finally note here that upon publication of this work, we will release all code publicly on GitHub.

% \tableofcontents

\section{Reinforcement Learning}
\label{sec:app_rl}
Reinforcement Learning (RL) is the problem of training machine learning models to solve sequential decision making problems. By interacting with an environment, RL agents must learn optimal behaviours given the current state of their environment. If the environment is fully observable, we can frame it as a Markov Decision Process (MDP) modelled as $\tuple{S, A, T, R, \gamma}$ where $S$ is the environment's state space, $A$ is the action space, $T(s_{t+1} | s_t, a_t)$ where $s_{t+1}, s_t \in S$ and $a_t \in A$ is the conditional transition probability between states $s_{t+1}$ and $s_t$ given action $a_t$, $r_t = R(s, a) : S \times A \rightarrow \mathbb{R}$ is the reward function for state action pair $(s, a)$, and $\gamma \in [0, 1]$ is the discount factor. The goal for an RL agent is to learn some optimal policy $\pi^*(a | s)$ that maximizes the expected discounted return $\expect_{\pi}{\left[\sum_{k=0}^{\infty}\gamma^kr_{t+k} \middle| S_t=s\right]}$. A single game is an \textit{episode}, and steps in an episode are indexed by $t$.

\section{Partially Observed Reinforcement Learning}
\label{sec:app_pomdp}
In a partially observed environment, an agent does not have access to the full state space $S$. We can frame this environment as a Partially Observable MDP (POMDP) modelled by $\tuple{S, A, T, O, \Omega, R, \gamma}$. In this new setting, $\tuple{S, A, T, R, \gamma}$ remain unchanged, $O$ represents the set of (partial) observations that the agent receives and $\Omega(o_t | s_t, a_{t-1})$ is the set of conditional observation probabilities.
An agent's goal is to learn some optimal policy $\pi^*(a | o)$ (or a policy that conditions on historical observations or on some internal memory) that maximizes the expected discounted return.

\section{TextWorld: Cooking Domain}
\label{sec:app_tw}
We present two examples of observations with instructions in \autoref{tab:instructions} and highlight where the instructions are and where the rewards come from.
\begin{table}[ht]
    \centering
        \caption{TextWorld observations for the Cooking Domain game. We show the observations and highlight where the instructions are, and finally identify what the rewards would be. This is for a level 2 game, and the total possible reward is $5$.}
    \begin{tabular}{p{0.95\textwidth}}
         \toprule
         \multicolumn{1}{c}{\textbf{Initial Game Observation}}\\
         \hdashline[.4pt/1pt]
         % array(["you are hungry ! let 's cook a delicious meal . check the cookbook in the kitchen for the recipe . once done , enjoy your meal ! -= kitchen = - you find yourself in a kitchen . you start to take note of what 's in the room . you can make out a closed fridge nearby . you can see an oven . you can make out a table . you wonder idly who left that here . you see a knife on the table . something scurries by right in the corner of your eye . probably nothing . you see a counter . the counter is vast . on the counter you see a raw yellow potato and a cookbook . you see a stove . but the thing is empty , unfortunately .",

         \texttt{``You are hungry! Let's cook a delicious meal. {\color{our_purple}Check the cookbook in the kitchen for the recipe.} {\color{our_maroon}Once done, enjoy your meal!}'' -=~kitchen~ =- you find yourself in a kitchen. You start to take note of what's in the room. You can make out a closed fridge nearby. You can see an oven. You can make out a table. You wonder idly who left that here. You see a knife on the table. Something scurries by right in the corner of your eye. Probably nothing. You see a counter. The counter is vast. On the counter you see a raw red potato and a cookbook. You see a stove, but the thing is empty, unfortunately.''}\\

         \midrule
         \multicolumn{1}{c}{\textbf{Reward}}\\
         \hdashline[.4pt/1pt]
         
         There is a reward of $1$ given for eating the meal. i.e. the instruction \texttt{``{\color{our_maroon}Once done, enjoy your meal!}''} will result in a reward of $1$ \textit{after} the recipe has been completed. Note that  the instruction \texttt{``{\color{our_purple}Check the cookbook in the kitchen for the recipe.}''} is not bound to a reward.\\
         
         \\
         \toprule
         
         \multicolumn{1}{c}{\textbf{Observation following the $\texttt{examine cookbook}$ action}}\\
         \hdashline[.4pt/1pt]
         
        \texttt{``You open the copy of \textit{``Cooking : a modern approach (3rd ed.)''}  and start reading: 
        recipe \#1 --------- {\color{our_green} Gather all following ingredients and follow the directions to prepare this tasty meal. Ingredients: red potato: directions: chop the red potato, fry the red potato, prepare meal}''}\\
        \midrule
         \multicolumn{1}{c}{\textbf{Reward}}\\
         \hdashline[.4pt/1pt]
         There are $4$ rewards from the instruction \texttt{``{\color{our_green} Gather all following ingredients and follow the directions to prepare this tasty meal. Ingredients: red potato: directions: chop the red potato, fry the red potato, prepare meal}''}:
        \begin{itemize}
            \item $1$ for grabbing the red potato  
            \item $1$ for chopping the red potato
            \item $1$ for frying the red potato
            \item $1$ for preparing the meal
        \end{itemize}\\
         \bottomrule
    \end{tabular}
    \label{tab:instructions}
\end{table}

\section{An example of LTL progression}
\label{sec:app_progression}
To illustrate how progression works, the LTL instruction $(\ltleventuallyTXT \ltlpredicate{player-has-carrot}) \wedge (\ltleventuallyTXT \ltlpredicate{player-has-apple})$ would be progressed to $(\ltleventuallyTXT \ltlpredicate{player-has-apple})$ once the agent grabs the carrot during an episode. In other words, when the agent reaches a state where $\ltlpredicate{player-has-carrot}$ is true, the LTL instruction is progressed to reflect that the agent no longer needs to get the carrot but must still grab the apple at some point.

\section{Generating LTL in TextWorld}
\label{sec:app_ltl_generation}
We provide some examples of the LTL instructions used in this work in \autoref{tab:ltl_gen_l3}, \autoref{tab:ltl_gen_l1}, and \autoref{tab:ltl_gen_l2}. We build a simple translator that reads game observations and constructs these LTL instructions directly, but only once. Repeated observations will not result in the same LTL formula being generated. Once a formula has been generated, LTL progression is used with the agent's belief state to progress the instructions along the truth assignments: observations are not directly used in the progression, although they do indirectly affect the progression by affecting the belief state.

For levels 0, 1, and 2, the LTL instructions that an agent can receive throughout an episode are (a) the task to examine the cookbook and (b) the recipe-bound task. In other words, the set of un-progressed instructions $%\varphi \in 
\Phi$ it 
% \revisit{\comment{can see}{TK: I guess that's not quite the right phrase, since it can also see progressed versions of those. Maybe ``receives''?} over the course of an episode (assuming the cookbook is examined)} 
can receive over the course of an episode (assuming the cookbook is examined)
is as follows:
\begin{align*}
    \Phi: [&\ltlnextTXT \ltlpredicate{cookbook-is-examined},\\
        &(\ltleventuallyTXT p_1) \wedge (\ltleventuallyTXT p_2) \wedge \hdots (\ltleventuallyTXT p_n)]
\end{align*}
where the recipe requires that predicates $p_1, p_2, \hdots p_n$ be true. Note that we also consider eating the meal to be a part of recipe in this case, although it is not explicitly mentioned in the recipe. Further, we note that the ``prepare meal'' task is represented by the predicate $\ltlpredicate{meal-in-player}$, as this is the event that occurs when the meal is prepared in the game.

For levels with navigation (i.e. level 3), 
\begin{align*}
    \Phi: [&\ltleventuallyTXT \ltlpredicate{player-at-kitchen},\\
          &\ltlnextTXT \ltlpredicate{cookbook-is-examined},\\
        &(\ltleventuallyTXT p_1) \wedge (\ltleventuallyTXT p_2) \wedge \hdots (\ltleventuallyTXT p_n)]
\end{align*}
where the agent has the added task of first navigating to the kitchen. This instruction provides no help for actually how to arrive at the kitchen, only that the agent must do so. As a result, \agentname\ still suffers from the difficulties of exploration, and perhaps investigating how LTL can be used to improve in navigation could be a direction for future work.

In total, LTL generation occurs only twice for any level, either during the initial observation or when the cookbook is read. When multiple instructions are generated at once, the agent will process them sequentially, in the order they are given.
% \commenttk{It's not clear to me how the agent deals with having two instructions generated at once at the start of level 3. I guess it somehow handles them in sequence, similar to how it would handle the single instruction $\ltleventuallyTXT(\ltlpredicate{player-at-kitchen}\wedge \ltlnextTXT \ltlpredicate{cookbook-is-examined})$?} \commentmt{Yes, it's in sequence. I tried modelling the instructions together at the start but ran into issues. For example, the formula " ltl = ('eventually', ('and', 'a', ('next', ('next', 'b'))))" errors when you try to run it with the progression code from ltl2action, which they took from Rodrigo i believe. I kept running into a few issues trying to combine next and eventually in the progression code initially, and never revisited trying to fix the progression code since I was not totally familiar with it. Ultimately feeding them in succession accomplishes the same thing, but it would have been nice to may have one big formula. I don't think performance really would change though, since really the transformer tends to focus on the beginning of the formula anyways, just based how the game plays out}
\begin{table}[ht]
    \centering
        \caption{Level 3 observation and resulting generated LTL instruction}

    \begin{tabular}{p{\textwidth}}
         \toprule
         \multicolumn{1}{c}{\textbf{Observation}}\\
         \hdashline[.4pt/1pt]
         \texttt{``You are hungry! Let's cook a delicious meal. Check the cookbook in the kitchen for the recipe. Once done, enjoy your meal!'' -=~corridor~=- ``You've entered a corridor. There is a closed screen door leading west. You don't like doors? Why not try going north, that entranceway is not blocked by one. You need an exit without a door? You should try going south.''}\\
        \midrule
         \multicolumn{1}{c}{\textbf{Generated LTL}}\\
         \hdashline[.4pt/1pt]
          This observation will generate two instructions: First,
          \begin{align*}
            \varphi: (\ltleventuallyTXT \ltlpredicate{player-at-kitchen})
            \end{align*}
            and second, 
            \begin{align*}
            \varphi: (\ltlnextTXT \ltlpredicate{cookbook-is-examined})
            \end{align*}\\
         \bottomrule
    \end{tabular}
    \label{tab:ltl_gen_l3}
\end{table}
\begin{table}[ht]
    \centering
        \caption{Level 1 observation and resulting generated LTL instruction}
    \begin{tabular}{p{\textwidth}}
         \toprule
         \textbf{Observation}\\
         \midrule
         \texttt{``You open the copy of \textit{``Cooking : a modern approach (3rd ed.)''}  and start reading: 
recipe \#1 --------- Gather all following ingredients and follow the directions to prepare this tasty meal. Ingredients: red potato: directions: chop the red potato, prepare meal''}\\
        \midrule
        \textbf{Generated LTL}\\
         \midrule
          {\begin{align*}
            \varphi: &(\ltleventuallyTXT \ltlpredicate{\scriptsize{red-potato-in-player}}) \wedge 
            (\ltleventuallyTXT \ltlpredicate{\scriptsize{red-potato-is-chopped}}) \wedge\\
             &(\ltleventuallyTXT \ltlpredicate{\scriptsize{meal-in-player}}) \wedge 
             (\ltleventuallyTXT \ltlpredicate{\scriptsize{meal-is-consumed}}).
            \end{align*}}\\
         \bottomrule
    \end{tabular}
    \label{tab:ltl_gen_l1}
\end{table}

\begin{table}[ht]
    \centering
        \caption{Level 2 observation and resulting generated LTL instruction}

    \begin{tabular}{p{\textwidth}}
         \toprule
         \textbf{Observation}\\
         \midrule
         \texttt{``You open the copy of \textit{``Cooking : a modern approach (3rd ed.)''}  and start reading: 
recipe \#1 --------- Gather all following ingredients and follow the directions to prepare this tasty meal. Ingredients: red potato: directions: chop the red potato, fry the red potato, prepare meal''}\\
        \midrule
         \textbf{Generated LTL}\\
         \midrule
          {\begin{align*}
            \varphi: &(\ltleventuallyTXT \ltlpredicate{\scriptsize{red-potato-in-player}}) \wedge 
            (\ltleventuallyTXT \ltlpredicate{\scriptsize{red-potato-is-chopped}}) \wedge\\
            &(\ltleventuallyTXT \ltlpredicate{\scriptsize{red-potato-is-fried}}) \wedge
             (\ltleventuallyTXT \ltlpredicate{\scriptsize{meal-in-player}}) \wedge \\
             &(\ltleventuallyTXT \ltlpredicate{\scriptsize{meal-is-consumed}}).
            \end{align*}}\\
         \bottomrule
    \end{tabular}
    \label{tab:ltl_gen_l2}
\end{table}
% you are hungry ! let 's cook a delicious meal . check the cookbook in the kitchen for the recipe . once done , enjoy your meal ! -= corridor = - you 've entered a corridor . there is a closed screen door leading west . you do n't like doors ? why not try going north , that entranceway is not blocked by one . you need an exit without a door ? you should try going south ."

\section{Model}
\label{sec:app_model}

\subsection{Text Encoder}
The text encoder is a simple transformer-based model, with a transformer block \citep{vaswani2017attention} and word embedding layer. We use the pre-trained 300-dimensional fastText \citep{mikolov2017advances} word embeddings, which are trained on Common Crawl (600B tokens). These word embeddings are frozen during training. Strings are tokenized by spaces.

The transformer block is composed of: \textbf{(1)} a stack of 5 convolutional layers, \textbf{(2)} a single-head self-attention layer, and \textbf{(3)} a 2-layer MLP with ReLU non-linear activation function in between. The convolutional layers each have 64 filters, with kernel sizes of 5 and are each followed by a Layer Norm \citep{ba2016layer}. We also use standard positional encoding \citep{vaswani2017attention}. The self-attention layer uses a hidden size $H$ of 64. The Text Encoder outputs a single feature vector $v \in \mathbb{R}^D$, where $D=64$ in our experiments. 

\subsection{Encoder Independence}
Figure \ref{fig:model} in the main paper visualizes each component of our model. Specifically, our model has four encoders: \textbf{(1)} Graph Encoder, \textbf{(2)} Text Encoder for observations, \textbf{(3)} Text Encoder for LTL instructions, and \textbf{(4)} Text Encoder for action choices. We note here that each of these encoders are independent models, trained concurrently. This is in contrast to the original GATA model that used the same Text Encoder for both the actions and the observations. Because these Text Encoders are relatively small transformers, there is no issues with fitting this model in memory. As shown in \autoref{tab:times}, the model is still quite efficient, even more than the original GATA code. We found that using independent encoders resulted in better performance than using a single Text Encoder that would have been responsible for encoding the observations, LTL instructions, and action choices.

\subsection{Action Selector}
The action selector is a simple two-layer MLP with a ReLU non-linear activation function in between. It takes as input, at time step $t$, the concatenated representation of the agent's state vector $z_t \in \mathbb{R}^{3D}$ and the action choices $C'_t \in \mathbb{R}^{n_c\times D}$. Recall that in our experiments $D=64$. The first layer uses an input dimension of $4D$ and an output dimension of $D$. The second layer has an input dimension of $D$ and output dimension of $1$, which after squeezing the last dimension during the forward pass, the final output vector $q_c \in \mathbb{R}^{n_c}$ represents the q-values for each action choice.

The input to the action selector is constructed by repeating the agent's state representation, $z_t$, $n_c$ times and then concatenating with the encoded actions choices $C'_t$. We wanted to further explain why this occurs, as it may not be immediately clear. The action selector in this work is a parameter-tied Q-value predictor. That is, for some action $a_i \in C_t,~i \in [1, \hdots, n_c]$ and agent state representation $z_t$, the predicted Q-value is $q_i = \texttt{AS}([a_i, z_t])$. Thus, the action selector (i.e. $\texttt{AS}(\cdot)$) predicts Q-values given action $a_i$ and agent state representation $z_t$. Thus, during a single episode step, given our encoded actions choices $C_t' \in \mathbb{R}^{n_c\times D}$, in order for the action selector to predict Q-values for each of these action choices, we repeat $z_t \in \mathbb{R}^{3D}$  $n_c$ times and stack it together, which results in a state matrix  $Z_t \in \mathbb{R}^{n_c \times 3D}$. When we concatenate this matrix with our action choices we are left with the input to our action selector: $[C'_t; Z_t] \in \mathbb{R}^{n_c \times 4D}$. Looking at this matrix, each row in this input matrix is effectively the concatenation of action $a_i$ with agent state representation $z_t$, and so passing this matrix to our action selector performs the parameter-tied Q-value prediction $q_i = \texttt{AS}([a_i, z_t])$ for all action choices, and outputs a single vector of Q-values for each action $q_c \in \mathbb{R}^{n_c}$. We can then use these predicted Q-values to perform action selection using either a greedy approach, an $\epsilon$-greedy approach, Boltzmann action selection, etc.

\section{Implementation Details}
\label{sec:imp_details}
\subsection{Augmenting GATA's Pre-Training Dataset}
We note here that although possible, the vocabulary and dataset used by \cite{adhikari2020learning} did not allow for the knowledge triple $\{cookbook, is, examined\}$ to be extracted from observations. Without this triple being extracted and added to the agent's belief state, there would be no way for the agent to progress LTL instructions requiring the agent to examine the cookbook. In our pre-training of the GATA graph encoder, we augmented the dataset provided by \cite{adhikari2020learning} to include the triplet $\{cookbook, is, examined\}$ when relevant (i.e. when the agent examines the cookbook). This was a simple process of adding this triple to the ground truth belief graphs in the dataset so that during pre-training, GATA could learn how to translate these triplets from relevant observations.

\subsection{Training}
\label{sec:app_training}
For training to learn our optimal policy we use the Double-DQN (DDQN) \citep{van2016deep} framework. We use $\epsilon$-greedy for training, which first starts with a warm-up period, using a completely random policy (i.e. $\epsilon=1.0$) for the first $1,000$ episodes. We then anneal $\epsilon$ from 1.0 to 0.1 over the next $3,000$ episodes after the initial warm-up (i.e. episodes $1,000$ to $4,000$). We use a prioritized experience replay buffer ($\alpha=0.6$ and $\beta=0.4$) with capacity $500,000$. For DDQN, the target network updates occur every $500$ episodes. We update network parameters every $50$ game steps, and we play $50$ games in parallel.

We train all agents for $100,000$ episodes using a discount factor of $0.9$, and we use $\{123, 321, 666\}$ as our random seeds. Each episode during training is limited to a maximum of $50$ steps, and during testing/validation this limit is increased to $100$ steps. We report results and save checkpoints every $1,000$ episodes. We also use a patience window $p$ that reloads from the previous best checkpoint during training when validation performance has decreased for $p$ episodes in a row. This is the same strategy used in \cite{adhikari2020learning}. For our experiments, we used $p=3$.

For reporting testing results, each model is trained using the three seeds mentioned before, and fine-tuned on the validation set. That is, the checkpoint of the model that performs best on the validation set during training is saved, and each of these models (three, one for each seed) is applied to the test set. Reported test results are the average over these three models.

\section{Experiments}
\label{sec:app_exp}
\subsection{Hyper-Parameters}
\label{sec:app_hp}
To have as fair a comparison as possible, we replicate all but three hyper-parameters from the settings used in \cite{adhikari2020learning}. We do this to remove any bias towards more finely tuned experimental configurations and focus only on the LTL integration. Further, we re-run the GATA experiments to confirm their original results. The three changes we implemented were (1) we use a batch size of 200 instead of 64 when training on the 100 game set, (2) for level 3, we use Boltzmann Action selection, and (3) we use Adam \citep{kingma2014adam} with a learning rate of $0.0003$ instead of RAdam \citep{DBLP:conf/iclr/LiuJHCLG020} with a learning rate of $0.001$. These changes boosted performance for all models. For the $20$ training game set, we use a batch size of $64$.

For Boltzmann action selection, we used a temperature of $\tau=100$. We experimented with various temperatures ($\tau \in \{1, 10, 25, 50, 100, 200\}$) and found $\tau=100$ to perform the best across models.

% \subsection{Following Instructions in GATA}
% \label{sec:app_stripped}
% Here we describe the experimental setup for the ablation study on GATA with instructions present in the observation versus with them them stripped. Note that 

\subsection{Computational Requirements}
\label{sec:comp_rec}
We report the wall-clock times for our experiments in \autoref{tab:times}.

% comment table out for now, seems to be bugging % TK: I don't think you can put \centering inside a tabular environment
\begin{table}[ht]
    \caption{Training times for each model and training set size. The times were reported using a workstation with dual RTX3090s, an AMD Ryzen 5950x 16-core CPU, and 128GB of RAM. For the graph updater, COC stands for the contrastive observation classification pre-training (the continuous belief graph model) and GTP stands for ground-truth pre-training (the discrete belief graph model).}

    \centering{
    \begin{tabular}{c|c|c|c}
         \toprule
         Model&Training Set Size&Batch Size&Approximate Time\\
         \midrule
         TDQN&20&64&16 hours\\
         LTL-GATA&20&64&24 hours\\
         GATA\textsuperscript{D}&20&64&24 hours\\
         GATA\textsuperscript{C}&20&64&24 hours\\
         $\text{GATA}^{\text{D}}_{\text{P}}$&20&64&36 hours\\
         $\text{GATA}^{\text{C}}_{\text{P}}$&20&64&36 hours\\
         \midrule
         TDQN&100&200&32 hours\\
         LTL-GATA&100&200&48 hours\\
         GATA\textsuperscript{D}*&100&200&48 hours\\
         GATA\textsuperscript{C}&100&200&48 hours\\
         $\text{GATA}^{\text{D}}_{\text{P}}$&100&200&65 hours\\
         $\text{GATA}^{\text{C}}_{\text{P}}$&100&200&65 hours\\
         \midrule
         \midrule
         Graph Updater using COC&N/A&64&48 hours\\
         Graph Updater using GTP&N/A&64&48 hours\\
         \bottomrule
    \end{tabular}
    \label{tab:times}
    }
\end{table}
% \subsection{Baseline Models}
% \label{sec:app_baselines}
% \textbf{Tr-DQN.}

% \textbf{GATA.}

% \subsection{Placeholder Header to Match Paper References}
% This section header is included to match references from the main paper, please ignore.
% \commenttk{I assume we can remove this for the camera-ready, since the paper references will also be updated.}

\subsection{Additional Results}
\label{sec:app_results}
\subsubsection{Ablation: Formatting LTL Predicates}
\label{sec:app_exp_pred}

\begin{figure}[ht]
    \centering
    \includegraphics[height=3.5cm]{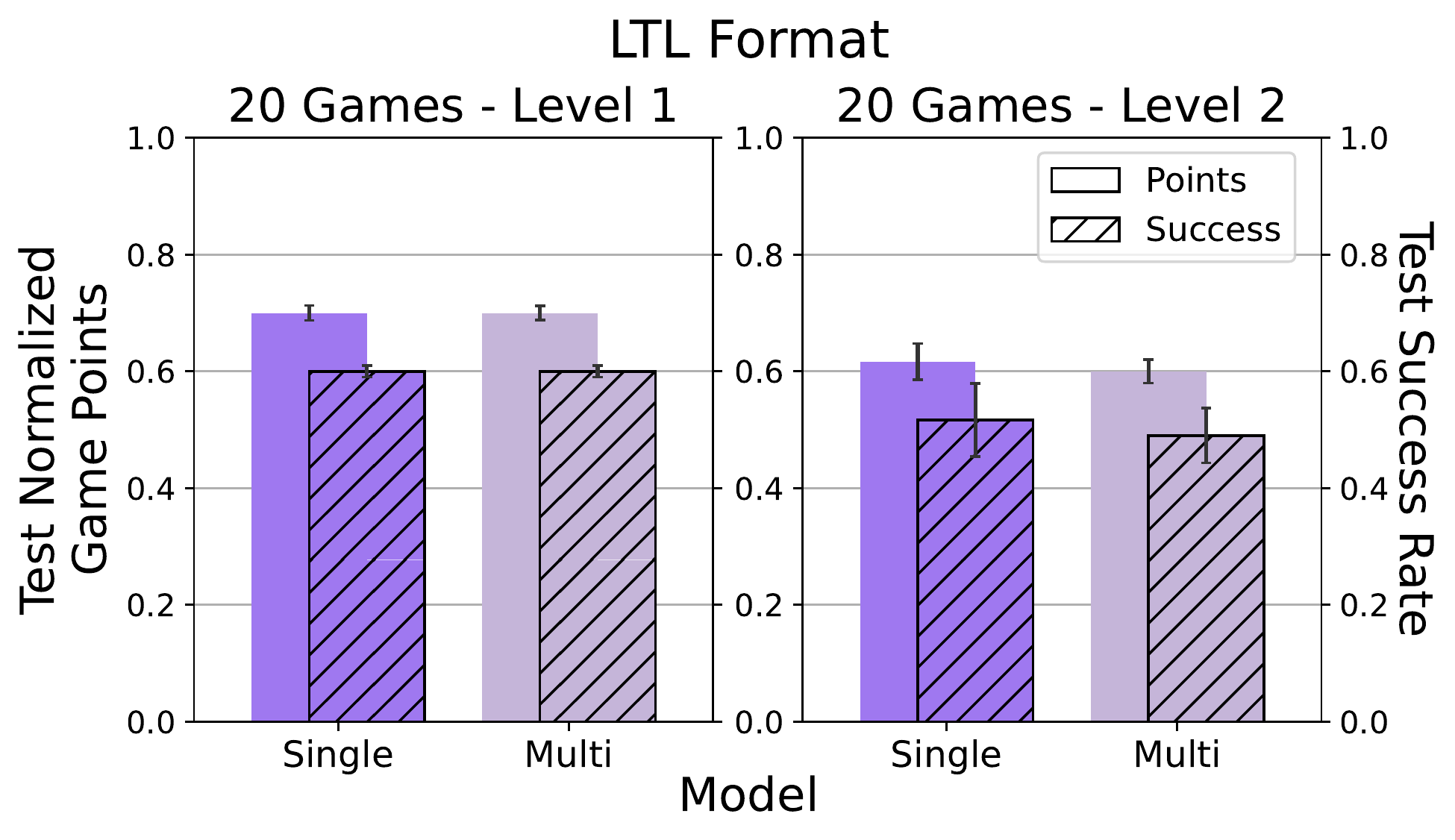}
    \caption{Study on LTL predicate format with single-token (\emph{Single}) predicates and multi-token (\emph{Multi}) predicates. Performance is largely unchanged with predicate format.}
    \label{fig:format_ablation}
\end{figure}
As we saw from \autoref{fig:level_perf}, \agentname\ when trained on the 100 games set performs significantly better than when trained on the 20 game set, which we attribute to the increased exposure to predicates during training, allowing it to generalize better during testing. To see if we can achieve the same level of generalization when training on the 20 game set, we compare \agentname\ with LTL predicates represented as single tokens (what we did in the main paper) with using multiple tokens. That is, we compare %\comment{\remove{against}}{TK: It's not ``against'' since we were already using one of them.} 
the following two string representations:
\begin{align*}
    \text{(single-token predicates)}~~&\text{str}(\varphi): \text{``eventually player\_has\_pepper and eventually pepper\_is\_cut''}\\
    \text{(multi-token predicates)}~~&\text{str}(\varphi): \text{``eventually player has pepper and eventually pepper is cut''}
\end{align*}
The single-token predicates are mapped in the vocabulary to a single word embedding. In our work, we compute word embedding for these single-token predicates by averaging the word embeddings of each underscore-separated word in the predicate. For example, the word embedding ($\texttt{WE}$) of the token $\ltlpredicate{player\_has\_pepper}$ is 
\begin{align*}
    \texttt{WE}(\ltlpredicate{player\_has\_pepper}) = \frac{\texttt{WE}(\ltlpredicate{player}) + \texttt{WE}(\ltlpredicate{has}) + \texttt{WE}(\ltlpredicate{pepper})}{3}
\end{align*}
For multiple-token predicates, each word has its own word embedding and we treat each word as any other word in the sentence. The idea is that by separating the tokens in the predicates, the text encoder (transformer) may be able to attend to each token independently, and during testing have better generalization. We visualize the results of this study in  \autoref{fig:format_ablation}. We can see from \autoref{fig:format_ablation} that this in fact does not help, and \agentname\ performs almost equally in either scenario. This does however show how our method is robust to predicate format.

% We relate these findings to similar findings by \cite{yao2021reading}, who showed that, RL frameworks like DDQN, randomly hashing the vector representations of the text-based observations and actions in text-based games results in equal and sometimes superior performance over methods that attempt to learn vector representations.

\subsubsection{The Effect of LTL Reward and LTL-Based Termination}
\label{sec:exp_reward_term}
\begin{figure}[t!]
    \centering
    \begin{tikzpicture}
      \node at (-3.6, 0){
        \includegraphics[height=3.5cm]{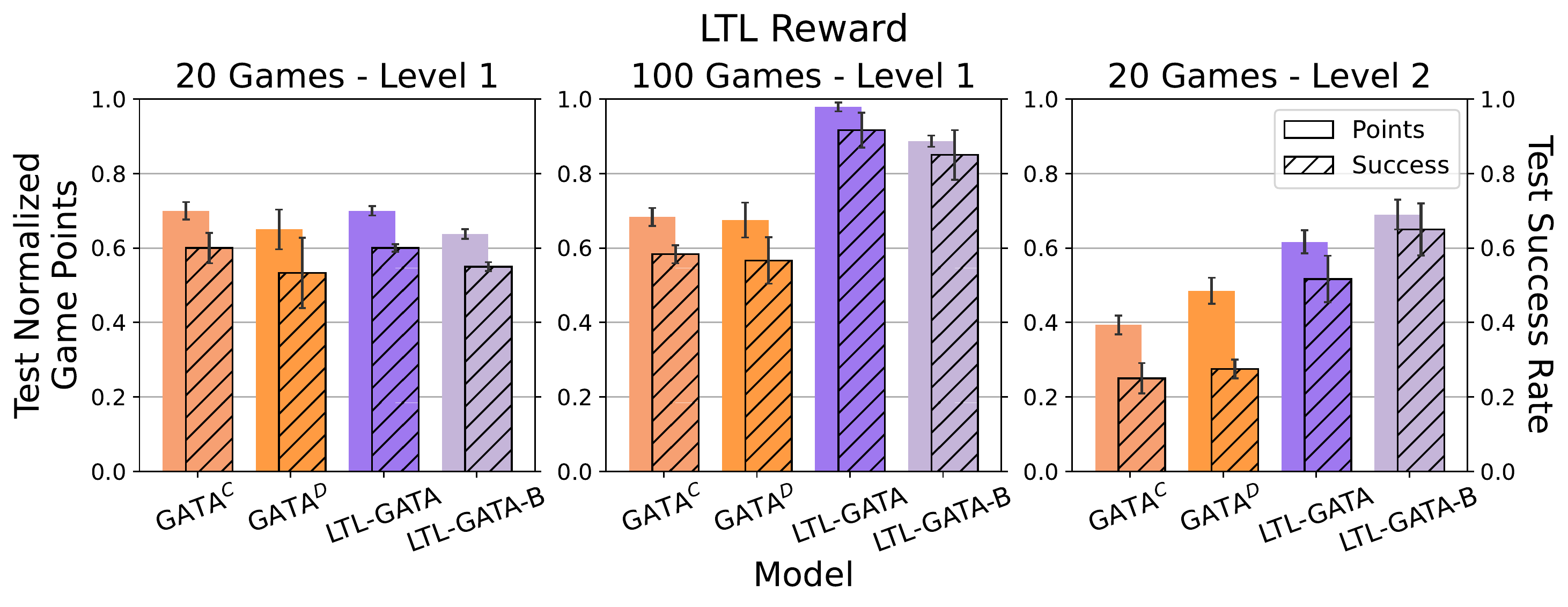}
      };
    %   \node at (3.4, 0){
    %     \includegraphics[height=3.3cm]{figures/test-bar-plots/ltl_progression_test_bars.pdf}
    %     };
      \node at (-3.6, 4){
        \includegraphics[height=3.5cm]{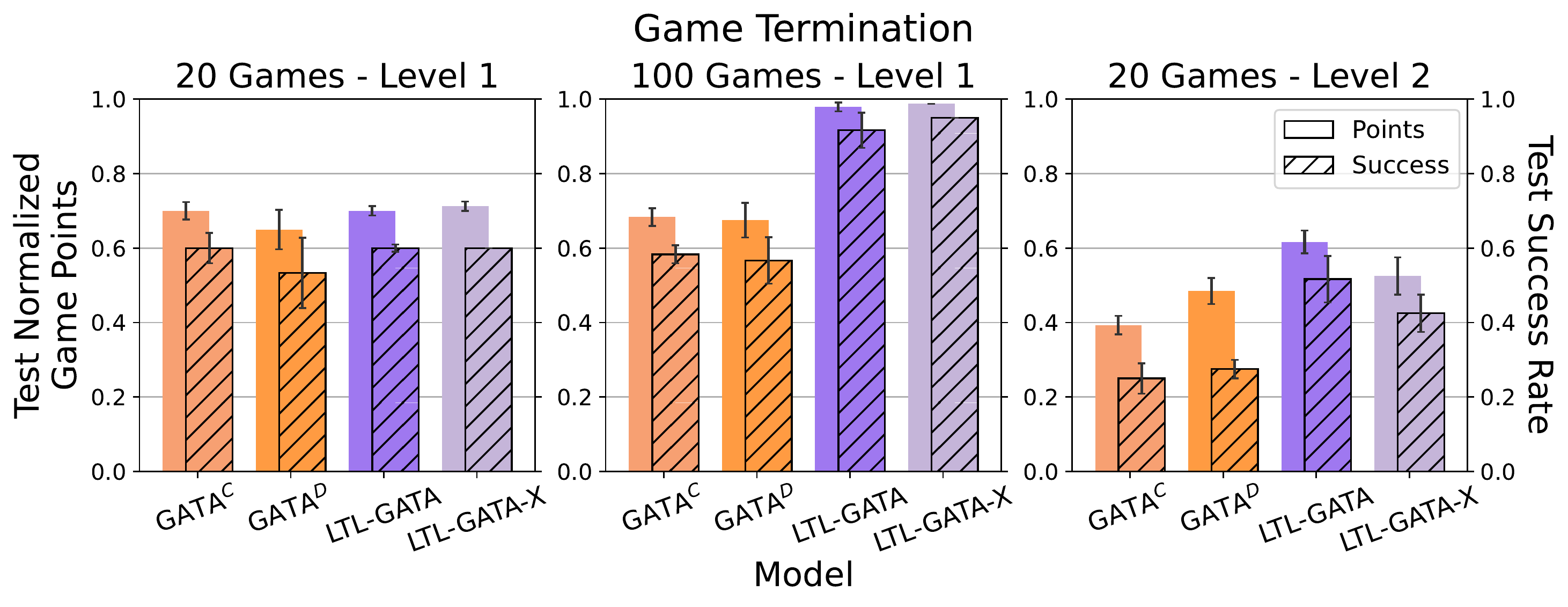}
      };
      \node at (3.4, 4){
          \includegraphics[height=3.5cm]{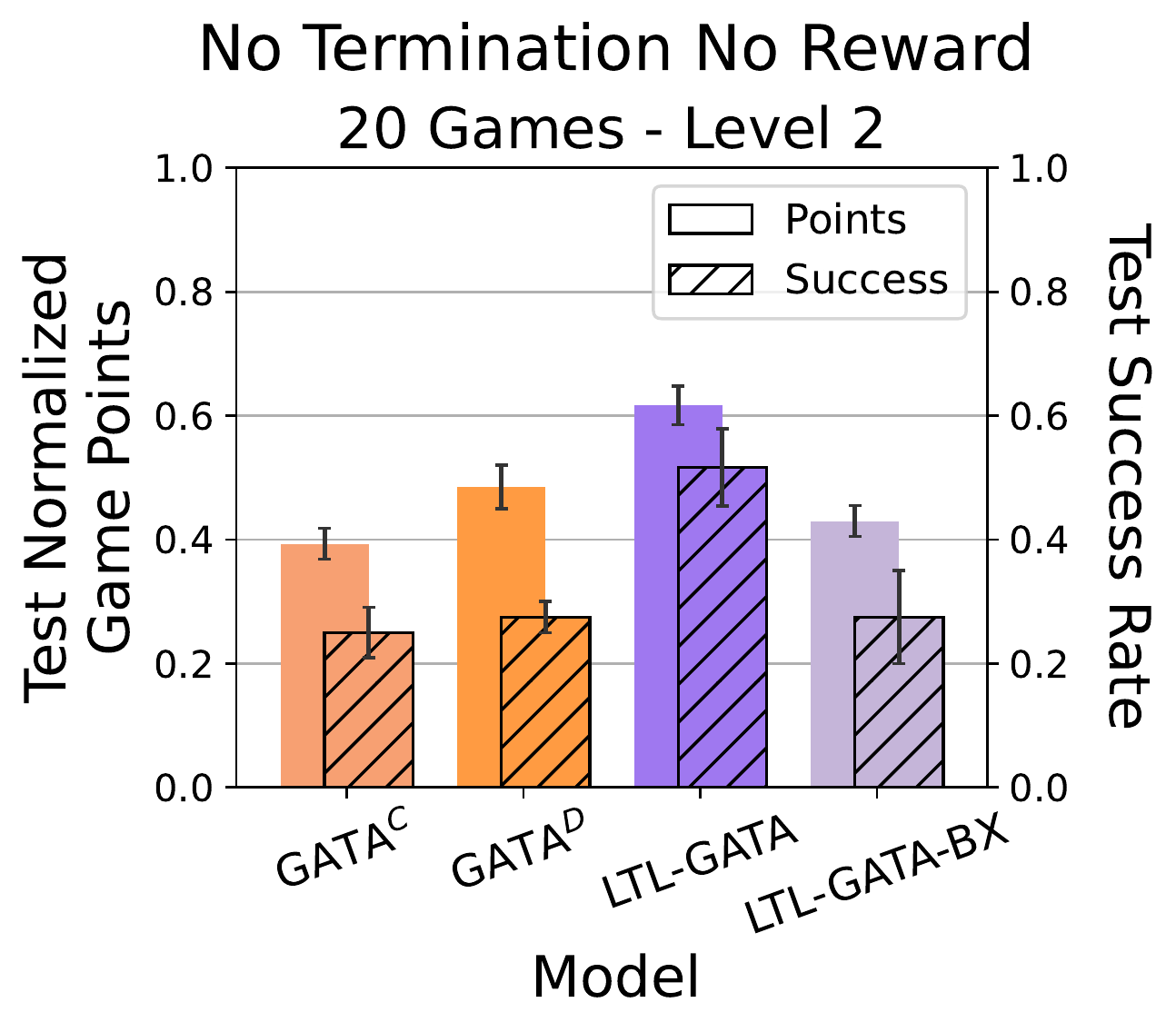}
        };
      \small
        \draw (-3.47, 2.05) node {(a)};
        \draw (-3.47, -2)  node {(b)};
        \draw (3.5, 2.05)  node {(c)};
        % \draw (3.5, -2)  node {(d)};
    \end{tikzpicture}

    \caption{Ablation studies on LTL-based episode termination and new reward function $\RLTL(s, a, \varphi)$ and on LTL progression. (a) LTL-GATA with new reward function $\RLTL(s, a, \varphi)$ and without LTL-based episode termination (\emph{LTL-GATA-X}). (b) LTL-GATA with base game reward function $R(s, a)$ and with LTL-based episode termination (\emph{LTL-GATA-B}). (c) LTL-GATA with base game reward function $R(s, a)$ and without LTL-based episode termination (\emph{LTL-GATA-BX}).}
    % \caption{Ablative studies on LTL-based episode termination and new reward function $\RLTL(s, a, \varphi)$. (a) \agentname\ with new reward function $\RLTL(s, a, \varphi)$ and without LTL-based episode termination. (b) \agentname\ with base game reward function $R(s, a)$ and with LTL-based episode termination. (c) \agentname\ with base game reward function $R(s, a)$ and without LTL-based episode termination.}
    \label{fig:reward_term}
\end{figure}
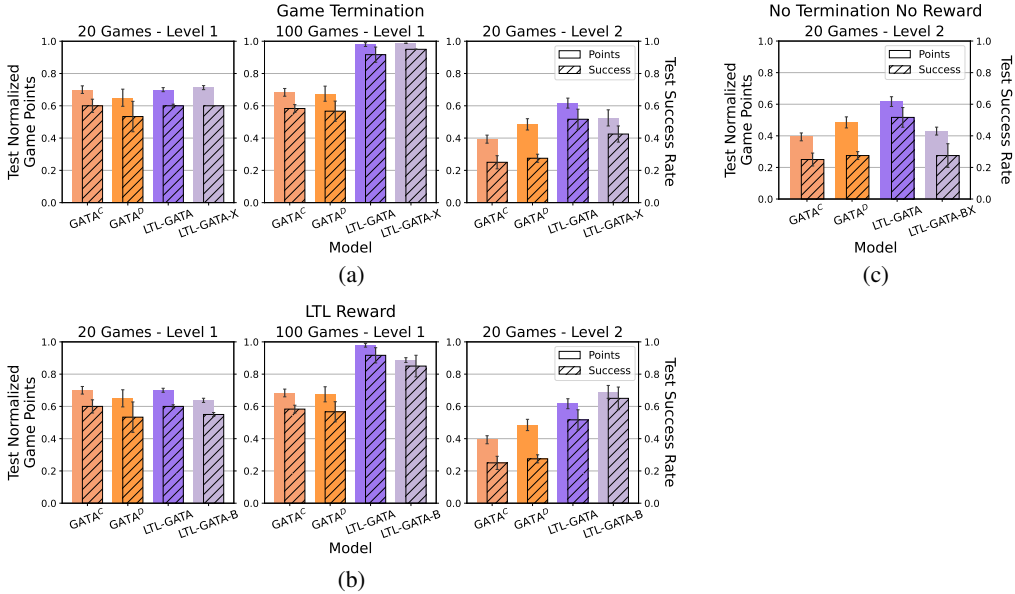

It is important to study the effect that the additional LTL bonus reward and LTL-based episode termination has on the performance of \agentname.  To study this, we consider three scenarios: \textbf{(a)} \agentname\ with the new reward function $\RLTL(s, a, \varphi)$ and without LTL-based episode termination; \textbf{(b)} \agentname\ with the base TextWorld reward function $R(s, a)$ and LTL-based episode termination; and \textbf{(c)} \agentname\ with the normal TextWorld reward function $R(s, a)$ and without LTL-based episode termination. For (a) and (b) we select level 1 on both the 20 and 100 game training set and level 2 on the 20 game training set. For (c) we select level 2 on 20 training games. We visualize the ablative study of these three scenarios in \autoref{fig:reward_term}. 

From \autoref{fig:reward_term} we can conclude that the presence of either the new reward function $\RLTL(s, a, \varphi)$ \textit{or} LTL-based episode termination is important to the performance of \agentname. This is because either of these methods will incentivize
 the agent to complete the initial $\ltlnextTXT \ltlpredicate{cookbook-is-examined}$ instruction, which isn't intrinsically rewarded by TextWorld. We can demonstrate the importance of this incentive by analyzing just one level (level 2 on 20 training games). Removing both methods leads to the agent not examining the cookbook, preventing it from receiving further instructions, which we can see from \autoref{fig:reward_term}(c) results in considerable performance loss, regressing to the baseline GATA. 

\subsection{Code}
All code for this work can be found at \codeurl.
% \commenttk{This should be updated to reflect that the code was put on GitHub (has it been yet?)}

\subsubsection{Fixing the GATA code}
\label{sec:app_fixed_gata}
We found two primary issues in the GATA code. First, we noticed that their implementation of the double Q-learning error was wrong. For Double Q-Learning, after performing some action $a_t$ in state $s_t$ and observing the immediate reward $r_{t}$ and resulting state $s_{t+1}$, the Q-Learning error is defined per \cite{van2016deep} as
\begin{align}
    Y_t = r_{t} + \gamma Q(s_{t+1}, \argmax_a Q(s_{t+1}, a; \boldsymbol{\theta}_t); \boldsymbol{\theta}'_t)
    \label{eq:err}
\end{align}
where $\boldsymbol{\theta}_t$ and $\boldsymbol{\theta}_t'$ are the parameters of the policy network and the target network, respectively. However, we noticed that the original code for GATA was computing the error as \footnote{\url{https://github.com/xingdi-eric-yuan/GATA-public/blob/c1afc3c9ab38256f839b3e0ddf8243796df5bd77/dqn_memory_priortized_replay_buffer.py\#L120-L123}}
\begin{align*}
    Y_t = r_{t} {\color{red}+ r_{t+1}} + \gamma Q(s_{t+1}, \argmax_a Q(s_{t+1}, a; \boldsymbol{\theta}_t); \boldsymbol{\theta}'_t)
\end{align*}
In other words, the reward for the stepped state was also being added to the error. 

Second, we found that the double Q-learning error for terminal states was being incorrectly implemented. Specifically, when computing the error for the case where $s_t$ is a terminal state, and therefore the stepped state $s_{t+1}$ does not exist, the stepped state was not being masked \footnote{\url{https://github.com/xingdi-eric-yuan/GATA-public/blob/c1afc3c9ab38256f839b3e0ddf8243796df5bd77/agent.py\#L1353-L1369}}. Additionally, presumably because of this initial error, terminal states were very rarely returned when sampling from experience, unless certain criteria were met \footnote{\url{https://github.com/xingdi-eric-yuan/GATA-public/blob/c1afc3c9ab38256f839b3e0ddf8243796df5bd77/dqn_memory_priortized_replay_buffer.py\#L93-L102}}. 

We found fixing these issues improved GATA’s performance considerably, which we demonstrated in \autoref{fig:level_perf}, and all our experimental results for GATA have this correction implemented.

\subsection{Training Curves}
Here we present accompanying training curves for experiments reported in this work. We report averaged curves of the normalized accumulated reward with bands representing the standard deviation.
\label{sec:app_train_curves}

\begin{figure}[h]
    \centering
    \includegraphics[height=3.5cm]{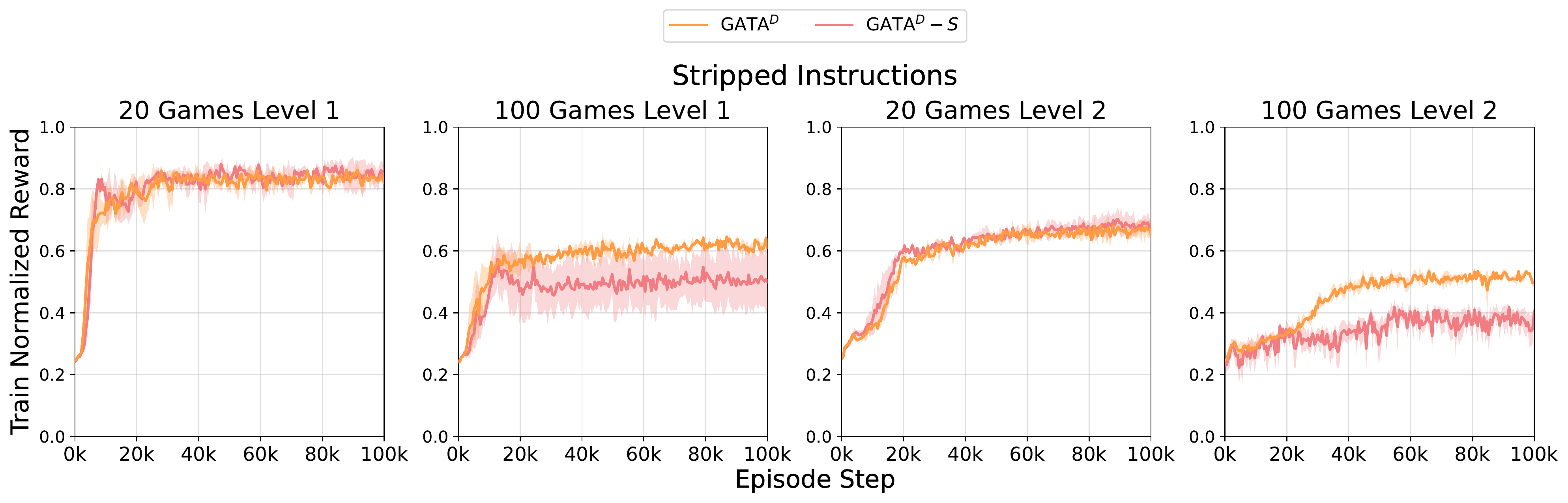}
\caption{Training curves (of normalized accumulated reward) for the comparison of GATA when trained with instructions (GATA\textsuperscript{D}) versus when instructions are stripped from environment observations (GATA\textsuperscript{D}-S). Agents were trained with 20 or 100 games, at increasing levels of task difficulty (level 1 vs level 2). Bands represent the standard deviation.}
\end{figure}

\begin{figure}[h]
    \centering
    \includegraphics[height=3.5cm]{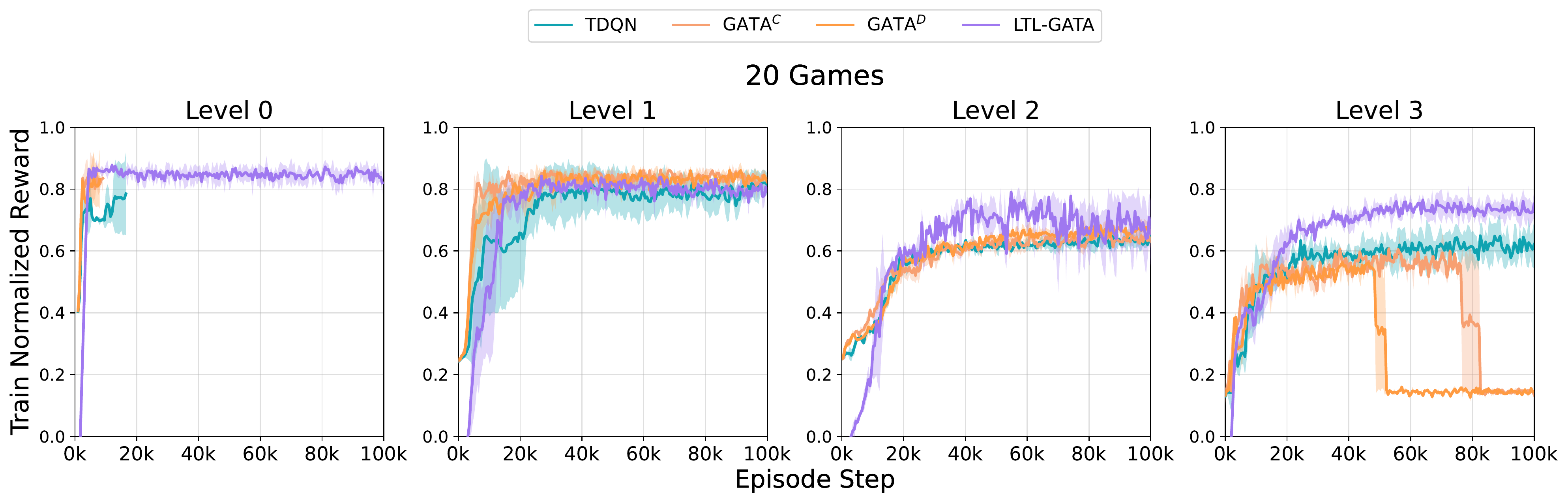}\\
    \includegraphics[height=3.5cm]{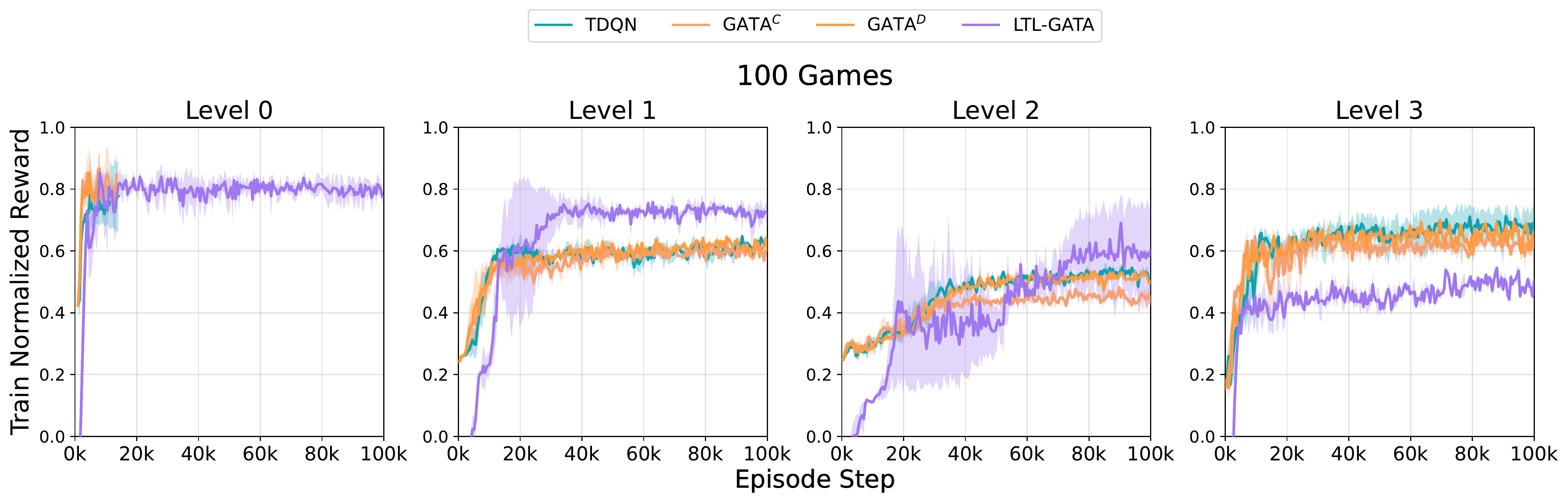}\\
\caption{Training curves of the normalized accumulated reward across various levels and on both the 20 (top) and 100 (bottom) game training sets. Bands represent the standard deviation. Note that on level 0, training curves for TDQN, GATA\textsuperscript{C}, and GATA\textsuperscript{D} were early stopped for achieving $\geq 0.95$ normalized accumulated reward on the validation set for $5$ episodes in a row.}
\end{figure}

\begin{figure}[h]
    \centering
    \includegraphics[height=3.3cm]{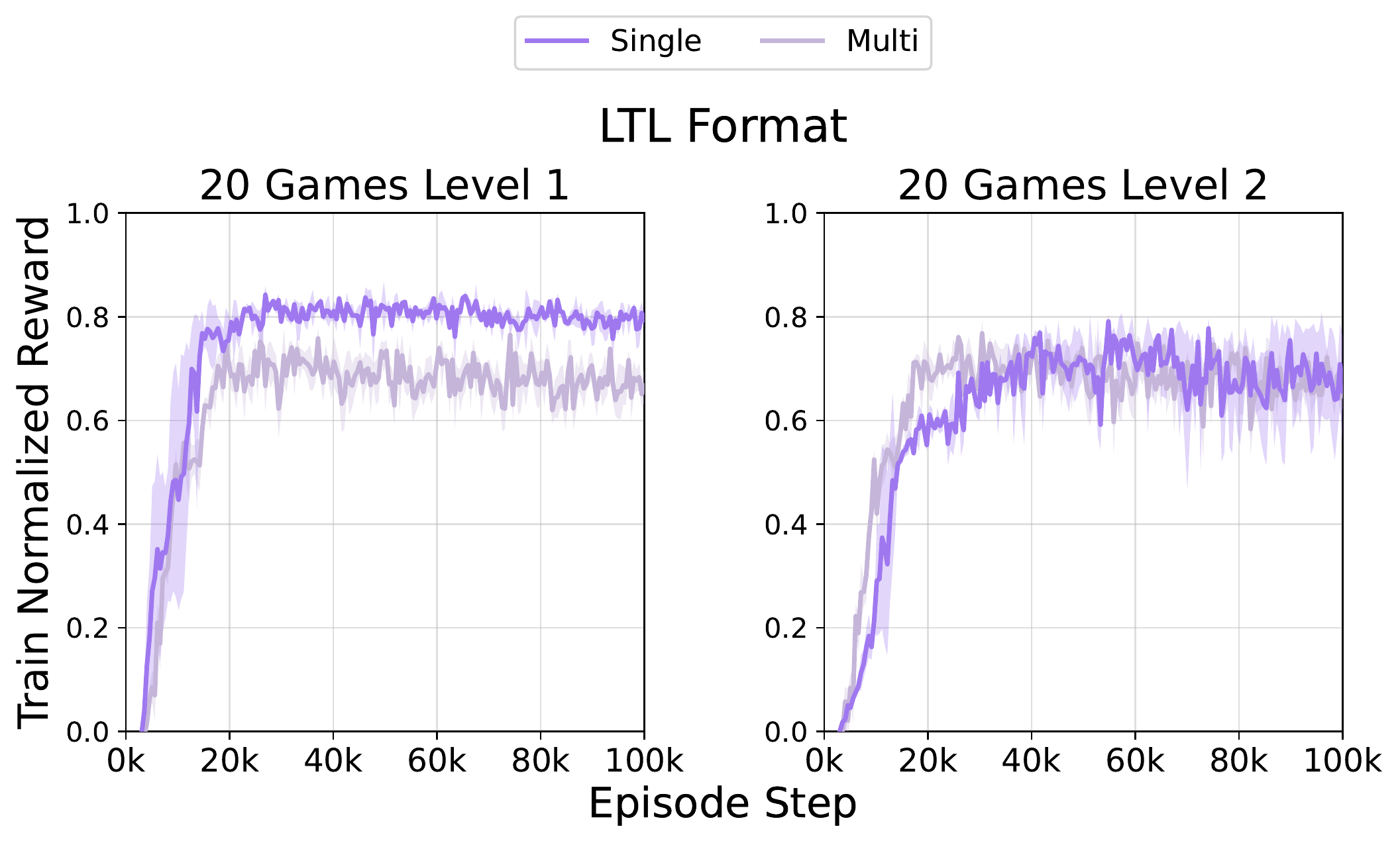}
    \caption{Training curves (of normalized accumulated reward) for the study on LTL predicate format with single-token (\emph{Single}) and multi-token (\emph{Multi}) predicates. Bands represent the standard deviation.}
\end{figure}

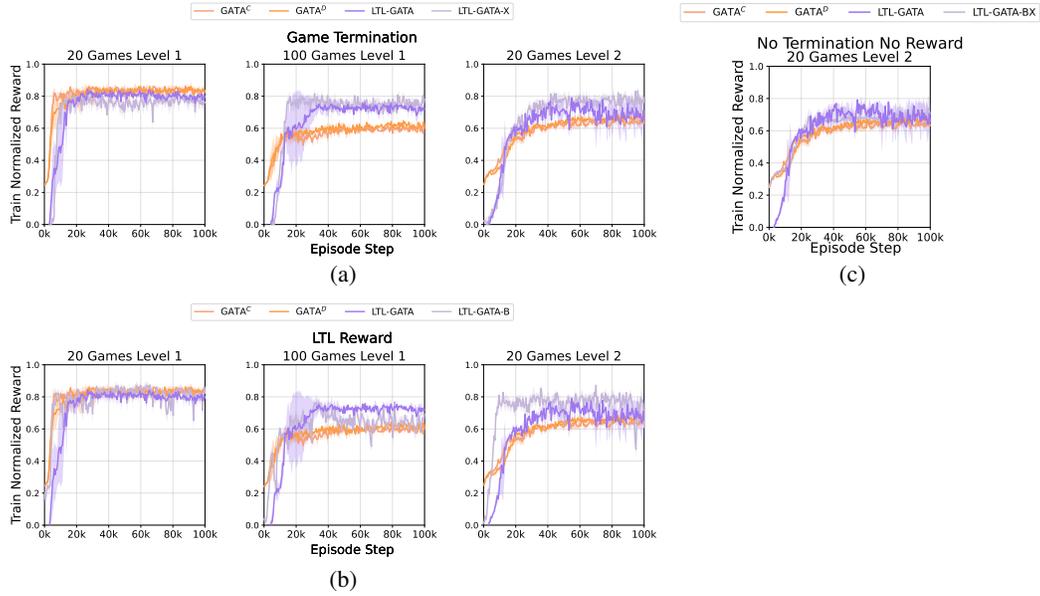
\begin{figure}[h]
    \centering
    \begin{tikzpicture}
      \node at (-3.6, 0){
        \includegraphics[height=3.5cm]{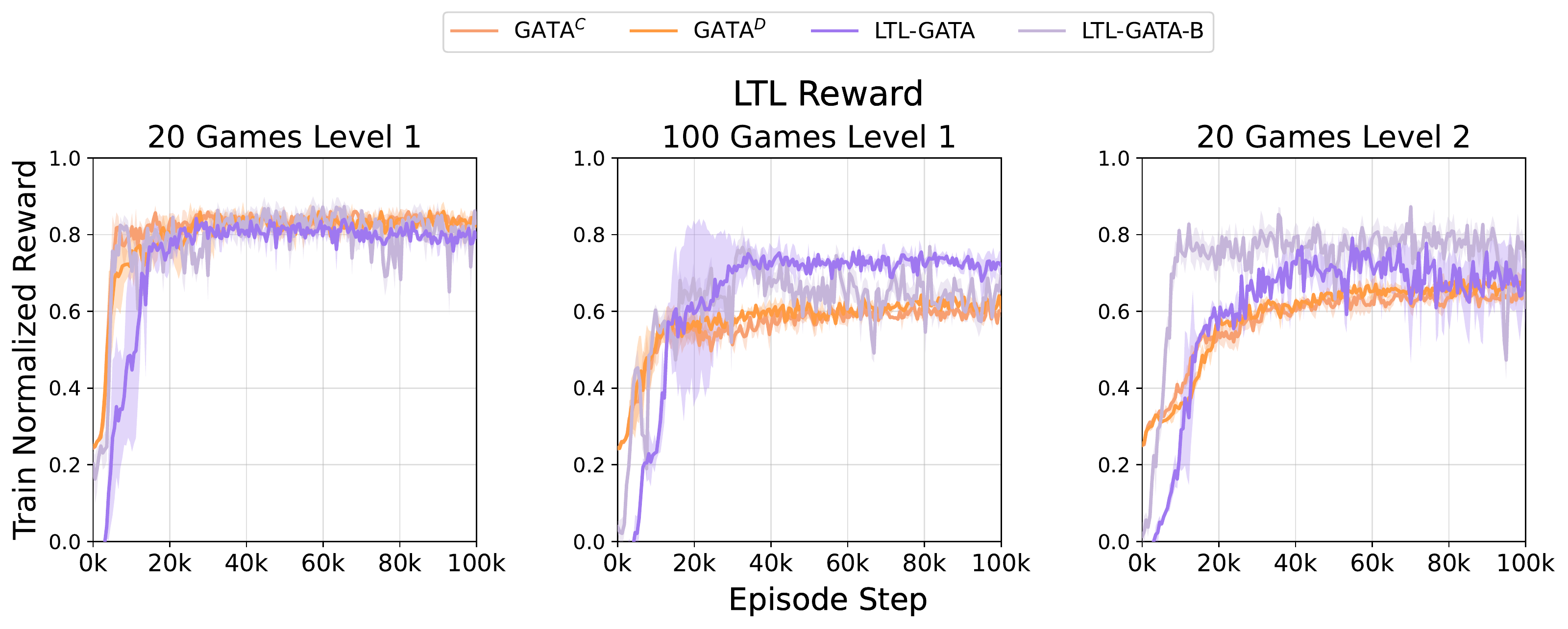}
      };
    %   \node at (3.4, 0){
    %     \includegraphics[height=3.3cm]{figures/test-bar-plots/ltl_progression_test_bars.pdf}
    %     };
      \node at (-3.6, 4){
        \includegraphics[height=3.5cm]{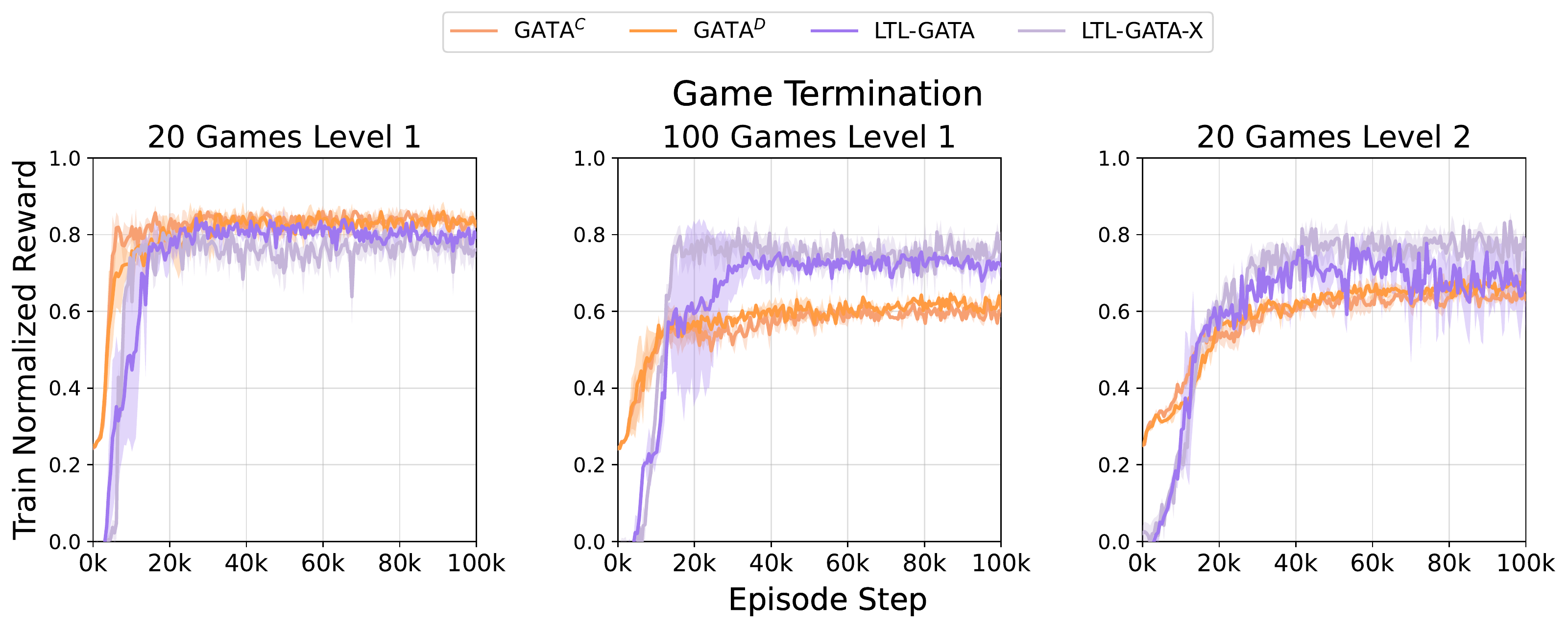}
      };
      \node at (3.4, 4){
          \includegraphics[height=3.5cm]{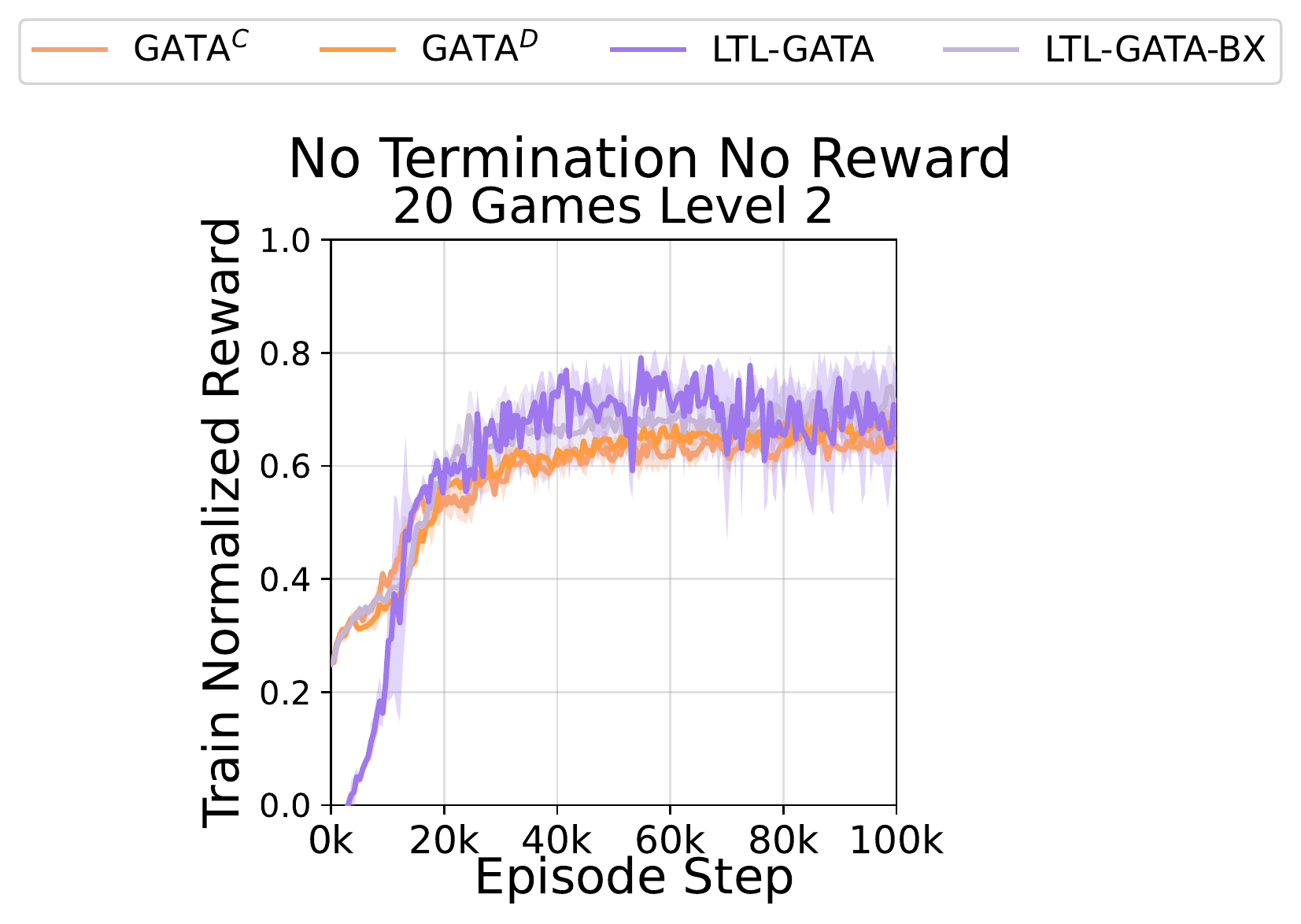}
        };
      \small
        \draw (-3.47, 2.05) node {(a)};
        \draw (-3.47, -2)  node {(b)};
        \draw (3.3, 2.05)  node {(c)};
        % \draw (3.5, -2)  node {(d)};
    \end{tikzpicture}

    \caption{Training curves (of normalized accumulated reward) for the ablation studies on LTL-based episode termination and new reward function $\RLTL(s, a, \varphi)$. (a) LTL-GATA with new reward function $\RLTL(s, a, \varphi)$ and without LTL-based episode termination (\emph{LTL-GATA-X}). (b) LTL-GATA with base game reward function $R(s, a)$ and with LTL-based episode termination (\emph{LTL-GATA-B}). (c) LTL-GATA with base game reward function $R(s, a)$ and without LTL-based episode termination (\emph{LTL-GATA-BX}). Bands represent the standard deviation.}
\end{figure}

\begin{figure}[h]
     \centering
    \begin{subfigure}{0.75\textwidth}
    \includegraphics[height=3.2cm]{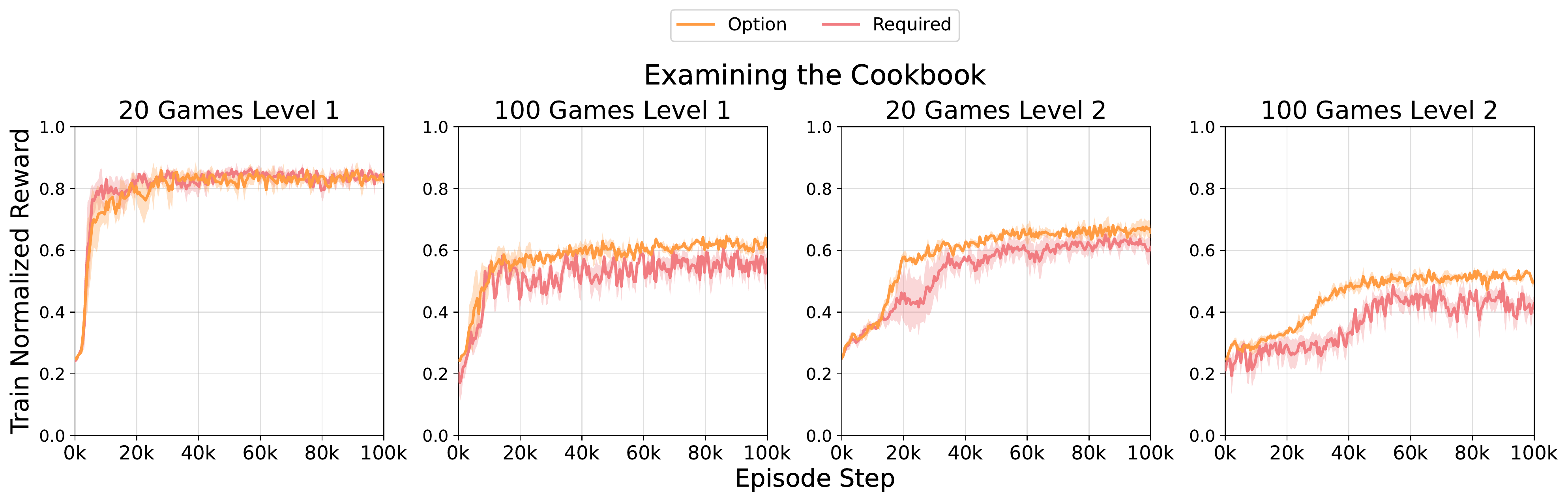}
    \caption{Letting GATA Consume the Cookbook}
    \end{subfigure}
    \begin{subfigure}{0.24\textwidth}
    \includegraphics[height=3.2cm]{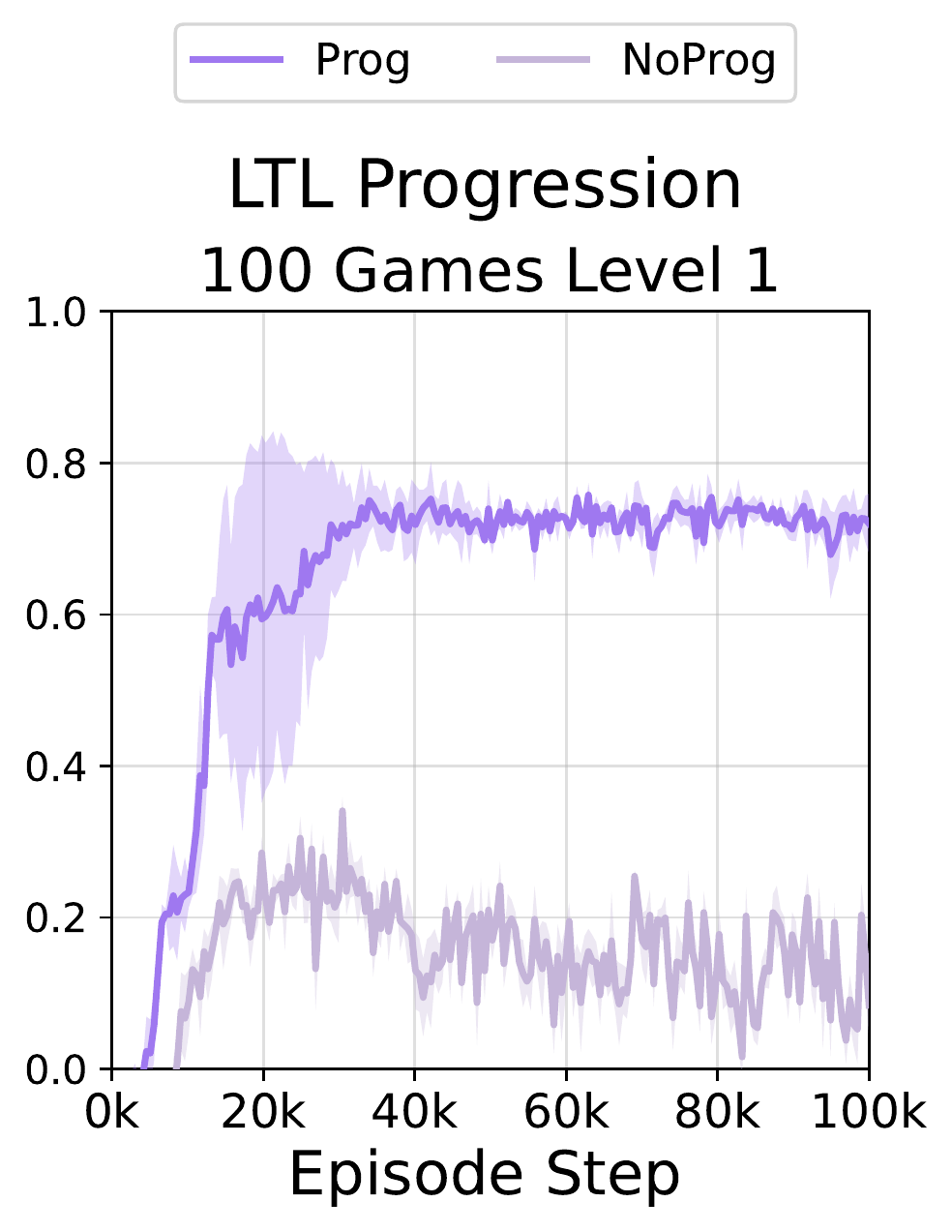}
    \caption{Progression Ablation}
    \end{subfigure}
    \caption{Training normalize reward curves for (a)  comparison of GATA\textsuperscript{D} when given the \emph{Option} to examine the cookbook vs. when it is \emph{Required} to examine the cookbook and (b)  comparison of LTL-GATA with (\emph{Prog}) and without (\emph{NoProg}) using LTL progression. Bands represent the standard deviation.}
    % \vspace{-5mm}
\end{figure}
\clearpage
\subsection{Automated Translation: Natural Language Instructions to LTL Details}
\label{sec:nl2ltldetails}

% Dataset details
% Prompt examples
% Example of DaVinci generalization
% Example of Ada failure

For the GPT-3 experiments on automated LTL translation in \autoref{sec:nl2ltl}, we simply extracted the observations used by our simple translator and saved the observation-translation pairs. Details of that translator were described in in \autoref{sec:app_ltl_generation} and examples of these pairs can be found in \autoref{tab:ltl_gen_l3} and \autoref{tab:ltl_gen_l1}. These observations are the sort of natural language we wish to translate into LTL, and we used six of the observation-translation pairs as the examples in our prompt to GPT-3.

% An example natural language/LTL pair as it appears in a prompt is shown below, with the natural language observation in turquoise and the translated LTL formula in red. 

The full prompt to GPT-3 is shown below (with colors added for readability). The six examples consist of a natural language observation (in turquoise) and a corresponding LTL formula (in red) --- these remain fixed for all prompts. The seventh line begins with the natural language observation to be translated (in blue). 

% \texttt{ 
% 1. {\color{our_green} NL: you open the copy of ``cooking : a modern approach ( 3rd ed . )'' and start reading : recipe \# 1 --------- gather all following ingredients and follow the directions to prepare this tasty meal . ingredients : cilantro directions : dice the cilantro prepare meal. } {\color{red}LTL: (‘and’, (‘eventually’, ‘cilantro\_in\_player’), (‘and’, (‘eventually’, ‘cilantro\_is\_diced’), (‘eventually’, ‘meal\_in\_player’)))}
% }\\

\texttt{
1. {\color{our_green} NL: you open the copy of ``cooking : a modern approach ( 3rd ed . )'' and start reading : recipe \# 1 --------- gather all following ingredients and follow the directions to prepare this tasty meal . ingredients : cilantro directions : dice the cilantro prepare meal
} {\color{red}LTL: (‘and’, (‘eventually’, ‘cilantro\_in\_player’), (‘and’, (‘eventually’, ‘cilantro\_is\_diced’), (‘eventually’, ‘meal\_in\_player’)))}
}

\texttt{
2. {\color{our_green}NL: you open the copy of ``cooking : a modern approach ( 3rd ed . )'' and start reading : recipe \# 1 --------- gather all following ingredients and follow the directions to prepare this tasty meal . ingredients : pork chop directions : chop the pork chop fry the pork chop prepare meal
} {\color{red}LTL: (‘and’, (‘eventually’, ‘pork\_chop\_in\_player’), (‘and’, (‘eventually’, ‘pork\_chop\_is\_chopped’), (‘and’, (‘eventually’, ‘pork\_chop\_is\_fried’), (‘eventually’, ‘meal\_in\_player’))))}
}

\texttt{
3. {\color{our_green}NL: you open the copy of ``cooking : a modern approach ( 3rd ed . )'' and start reading : recipe \# 1 --------- gather all following ingredients and follow the directions to prepare this tasty meal . ingredients : black pepper directions : prepare meal
} {\color{red}LTL: (‘and’, (‘eventually’, ‘black\_pepper\_in\_player’), (‘eventually’, ‘meal\_in\_player’))}}

\texttt{
4. {\color{our_green}NL: you open the copy of ``cooking : a modern approach ( 3rd ed . )'' and start reading : recipe \# 1 --------- gather all following ingredients and follow the directions to prepare this tasty meal . ingredients : purple potato red onion salt directions : dice the purple potato roast the purple potato dice the red onion fry the red onion prepare meal
} {\color{red}LTL: (‘and’, (‘eventually’, ‘purple\_potato\_in\_player’), (‘and’, (‘eventually’, ‘red\_onion\_in\_player’), (‘and’, (‘eventually’, ‘salt\_in\_player’), (‘and’, (‘eventually’, ‘purple\_potato\_is\_diced’), (‘and’, (‘eventually’, ‘purple\_potato\_is\_roasted’), (‘and’, (‘eventually’, ‘red\_onion\_is\_diced’), (‘and’, (‘eventually’, ‘red\_onion\_is\_fried’), (‘eventually’, ‘meal\_in\_player’))))))))} }

\texttt{
5. {\color{our_green}NL: you open the copy of ``cooking : a modern approach ( 3rd ed . )'' and start reading : recipe \# 1 --------- gather all following ingredients and follow the directions to prepare this tasty meal . ingredients : black pepper parsley salt directions : dice the parsley prepare meal
} {\color{red}LTL: (‘and’, (‘eventually’, ‘black\_pepper\_in\_player’), (‘and’, (‘eventually’, ‘parsley\_in\_player’), (‘and’, (‘eventually’, ‘salt\_in\_player’), (‘and’, (‘eventually’, ‘parsley\_is\_diced’), (‘eventually’, ‘meal\_in\_player’)))))}
}

\texttt{
6. {\color{our_green}NL: you open the copy of ``cooking : a modern approach ( 3rd ed . )'' and start reading : recipe \# 1 --------- gather all following ingredients and follow the directions to prepare this tasty meal . ingredients : purple potato white onion yellow bell pepper directions : roast the purple potato roast the white onion dice the yellow bell pepper prepare meal
} {\color{red}LTL: (‘and’, (‘eventually’, ‘purple\_potato\_in\_player’), (‘and’, (‘eventually’, ‘white\_onion\_in\_player’), (‘and’, (‘eventually’, ‘yellow\_bell\_pepper\_in\_player’), (‘and’, (‘eventually’, ‘purple\_potato\_is\_roasted’), (‘and’, (‘eventually’, ‘white\_onion\_is\_roasted’), (‘and’, (‘eventually’, ‘yellow\_bell\_pepper\_is\_diced’), (‘eventually’, ‘meal\_in\_player’)))))))} }

\texttt{
7. {\color{blue}NL: you open the copy of ``cooking : a modern approach ( 3rd ed . )'' and start reading : recipe \# 1 --------- gather all following ingredients and follow the directions to prepare this tasty meal . ingredients : banana red hot pepper yellow potato directions : chop the banana fry the banana chop the red hot pepper fry the red hot pepper slice the yellow potato fry the yellow potato prepare meal
} {\color{red}{LTL:}
}}

In \autoref{tab:gpt-3}, we show how \emph{Da Vinci} and \emph{Ada} complete the prompt to translate this example of a natural language TextWorld observation.
While \emph{Da Vinci} produces the correct LTL translation, \emph{Ada} struggles and hallucinates incorrect predicates such as \texttt{`zucchini\_fry\_player'} and \texttt{`banana\_is\_frozen'}.

\begin{table}[ht]
    \centering
        \caption{GPT-3 Natural-Language-to-LTL Translation Example}

    \begin{tabular}{p{\textwidth}}
         \toprule
         \multicolumn{1}{c}{\textbf{Observation}}\\
         \hdashline[.4pt/1pt]
            \texttt{ \color{blue}
                you open the copy of ``cooking : a modern approach ( 3rd ed . )'' and start reading : recipe \# 1 --------- gather all following ingredients and follow the directions to prepare this tasty meal . ingredients : banana red hot pepper yellow potato directions : chop the banana fry the banana chop the red hot pepper fry the red hot pepper slice the yellow potato fry the yellow potato prepare meal
            }\\
        \midrule
         \multicolumn{1}{c}{\textbf{\emph{Da Vinci} LTL Translation (\emph{Absolutely Correct})}}\\
         \hdashline[.4pt/1pt]
            \texttt{
          (‘and’, (‘eventually’, ‘banana\_in\_player’), (‘and’, (‘eventually’, ‘red\_hot\_pepper\_in\_player’), (‘and’, (‘eventually’, ‘yellow\_potato\_in\_player’), (‘and’, (‘eventually’, ‘banana\_is\_chopped’), (‘and’, (‘eventually’, ‘banana\_is\_fried’), (‘and’, (‘eventually’, ‘red\_hot\_pepper\_is\_chopped’), (‘and’, (‘eventually’, ‘red\_hot\_pepper\_is\_fried’), (‘and’, (‘eventually’, ‘yellow\_potato\_is\_sliced’), (‘and’, (‘eventually’, ‘yellow\_potato\_is\_fried’), (‘eventually’, ‘meal\_in\_player’))))))))))}\\
         
         \midrule
         
         \multicolumn{1}{c}{\textbf{\emph{Ada} LTL Translation \emph{(Incorrect)}}}\\
         \hdashline[.4pt/1pt]
            \texttt{
          (‘and’, (‘eventually’, ‘banana\_in\_player’), (‘and’, (‘eventually’, ‘red\_hot\_pepper\_in\_player’), (‘and’, (‘eventually’, ‘yellow\_potato\_in\_player’), (‘and’, (‘eventually’, ‘zucchini\_fry\_player’), (‘and’, (‘eventually’, ‘banana\_is\_frozen’), (‘eventually’, ‘meal\_in\_player’))))
          } \\
         \bottomrule

    \end{tabular}
    \label{tab:gpt-3}
\end{table}

\clearpage
\section{Broader Impact}
\label{sec:broader}
As \cite{adhikari2020learning} suggested, text-based games can be a proxy for studying human-machine interaction through language. Human-machine interaction and relevant systems have many potential ethical, social, and safety concerns. Providing inaccurate policies or information or partially completing tasks in critical systems can have devastating consequences. For example, in health care, improper treatment can be fatal, or in travel planning, poor interactions can lose a client money.

\citet[section 7]{adhikari2020learning} identified several research objectives relating to language-based agents:
improve the ability to make better decisions, allow for constraining decisions for safety purposes, and improve interpretability. We highlight how RL agents equipped with LTL instructions can improve in these areas.
For constraining decisions, it may be desirable to do so in way that depends on the history, which LTL gives a way to keep track of.
With respect to interpretability, we propose that monitoring the progression of instructions provides a mechanism for understanding where and when an agent might be making incorrect decisions, and provides the opportunity to revise instructions or attempt to fix the problem by other means.

However, instruction following, especially overly literal instruction following,
% \remove{or more specifically} \added{especially} \revisit{\remove{too} \added{overly} literal instruction following} \commentmt{need to reword}, 
may not always be beneficial and can even be harmful. \cite{DBLP:journals/corr/abs-2205-01975} describe a good example where an agent in the Zork1 game breaks into a home and steals the items it needs. In that specific case, breaking into the home has no adverse effect on the agent's reward, and so it has no %intrinsic 
incentive not to perform this act. %Although breaking into someone's home is an extreme (and perhaps even illegal) example, 
Violation of social norms like this are not modelled in our work, and can have negative impacts, even in less extreme cases. Furthermore, there are potential dangers of incorrect, immoral,  
%\commenttk{incorrect? immoral?}
or even misinterpreted instructions that lead to dangerous outcomes. 
% \comment{\added{In systems where an agent may interact with a human user, there may even be} instructions for the agent that are inconsiderate to the user (and their preferences) they may be trying to serve.}{TK: I'm not sure this adds anything on the previous sentence.} 
Although we do not directly address these concerns in this work, they pose interesting directions for future work.

\renewcommand\refname{Additional References}

\end{document}